\documentclass[acmsmall,authorversion]{acmart}
\AtBeginDocument{%
  }

\setcopyright{acmlicensed}
\acmJournal{PACMCGIT}
\acmYear{2025} \acmVolume{8} \acmNumber{1} \acmArticle{18} \acmMonth{5}\acmDOI{10.1145/3728302}

\usepackage{delarray}
\usepackage{multirow}
\usepackage{xcolor}
\usepackage{rotating}
\usepackage{amsmath}
\usepackage{subcaption}

\graphicspath{{figures/}{pictures/}{images/}{./}} %
\title{VR-Splatting: Foveated Radiance Field Rendering via 3D Gaussian Splatting and Neural Points}

\begin{document}
\author{Linus Franke}
\email{linus.franke@fau.de}
\orcid{0000-0001-8180-0963}
\author{Laura Fink}
\email{laura.fink@fau.de}
\orcid{0009-0007-8950-1790}
\author{Marc Stamminger}
\email{marc.stamminger@fau.de}
\orcid{0000-0001-8699-3442}
\affiliation{%
  \institution{Friedrich-Alexander-Universit\"at Erlangen-N\"urnberg}
  \country{Germany}
}

\begin{teaserfigure}
  \centering
  \includegraphics[width=\linewidth]{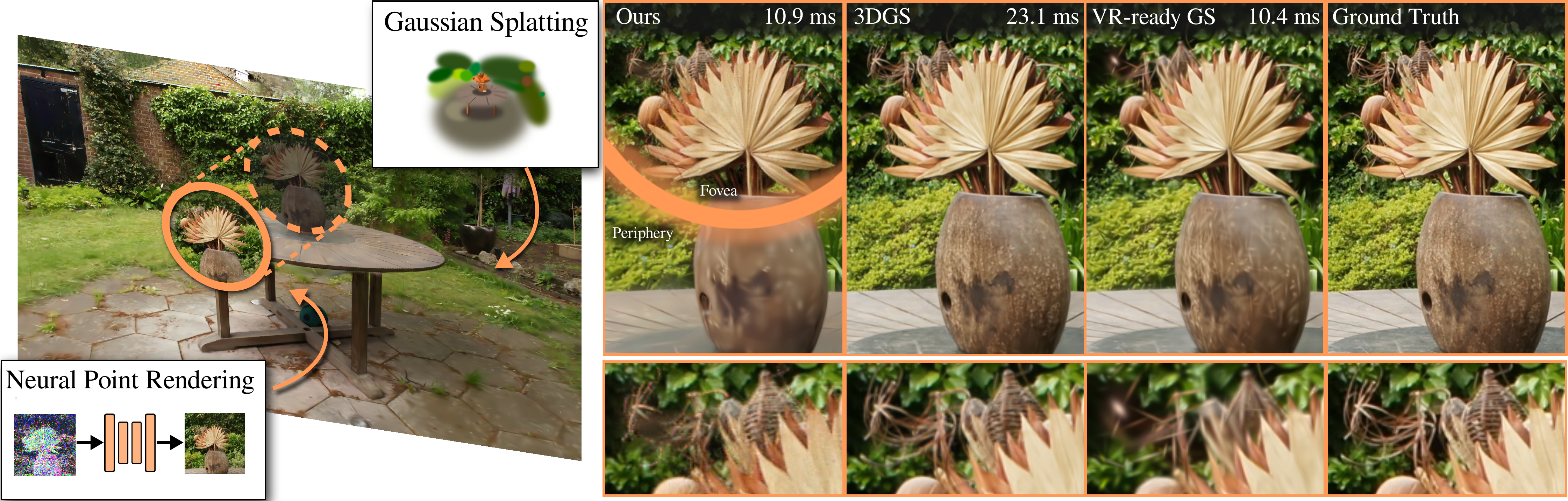}
  \caption{VR-Splatting combines advantages of neural point rendering~\cite{franke2024trips} and 3D Gaussian Splatting~(3DGS)~\cite{kerbl3Dgaussians} within a hybrid foveated radiance field rendering system for VR. In contrast to 3DGS and a VR-ready variant, which only uses 0.5 Mio. primitives (see Fig.~\ref{fig:motivation_plot}), our method provides high fidelity in the foveal region and meets the 90\,Hz VR framerate requirements at $2016\times2240$ pixels per eye. }
  \label{fig:teaser}
\end{teaserfigure}

\begin{abstract}
    Recent advances in novel view synthesis have demonstrated impressive results in fast photorealistic scene rendering through differentiable point rendering, either via Gaussian Splatting~(3DGS)~\cite{kerbl3Dgaussians} or neural point rendering~\cite{aliev2020npbg}. 
    Unfortunately, these directions require either a large number of small Gaussians or expensive per-pixel post-processing for reconstructing fine details, which negatively impacts rendering performance.
    To meet the high performance demands of virtual reality~(VR) systems, primitive or pixel counts therefore must be kept low, affecting visual quality.

    In this paper, we propose a novel hybrid approach based on foveated rendering as a promising solution that combines the strengths of both point rendering directions regarding performance sweet spots.
    Analyzing the compatibility with the human visual system, we find that using a low-detailed, few primitive smooth Gaussian representation for the periphery is cheap to compute and meets the perceptual demands of peripheral vision.    
    For the fovea only, we use neural points with a convolutional neural network for the small pixel footprint, which provides sharp, detailed output within the rendering budget.
    This combination also allows for synergistic method accelerations with point occlusion culling and reducing the demands on the neural network.
    
    Our evaluation confirms that our approach increases sharpness and details compared to a standard VR-ready 3DGS configuration, and participants of a user study overwhelmingly preferred our method.
    Our system meets the necessary performance requirements for real-time VR interactions, ultimately enhancing the user's immersive experience.
    The project page can be found at:
    \url{https://lfranke.github.io/vr_splatting}
\end{abstract} %

\maketitle

\section{Introduction}

Virtual teleportation into real-world environments, displayed on a virtual reality (VR) headset is a long-standing goal in computer graphics and computer vision. 
State-of-the-art VR headsets exhibit high-resolution and high-refresh-rate displays, which place high and strict performance demands on rendering algorithms.

In light of this, we propose a novel radiance field rendering method that allows interactive virtual scene exploration.
We meet the high performance demands by exploiting the visual acuity falloff in the periphery through \textit{foveated rendering}.
Most foveated rendering algorithms render the foveal region as sharp and crisp as possible while trying to cut computational costs in the periphery by, e.g., lowering the resolution~\cite{guenter2012foveated} or the number of rays traced~\cite{weier2016foveated}.
However, the application of such techniques is not trivial as they cause side effects such as temporal instability or flickering, to which the peripheral vision is highly sensitive~\cite{weier2017perception}.

\begin{figure}[b]
    \centering
    \begin{subfigure}[]{.48\linewidth}
    \centering
    \includegraphics[width=.99\linewidth]{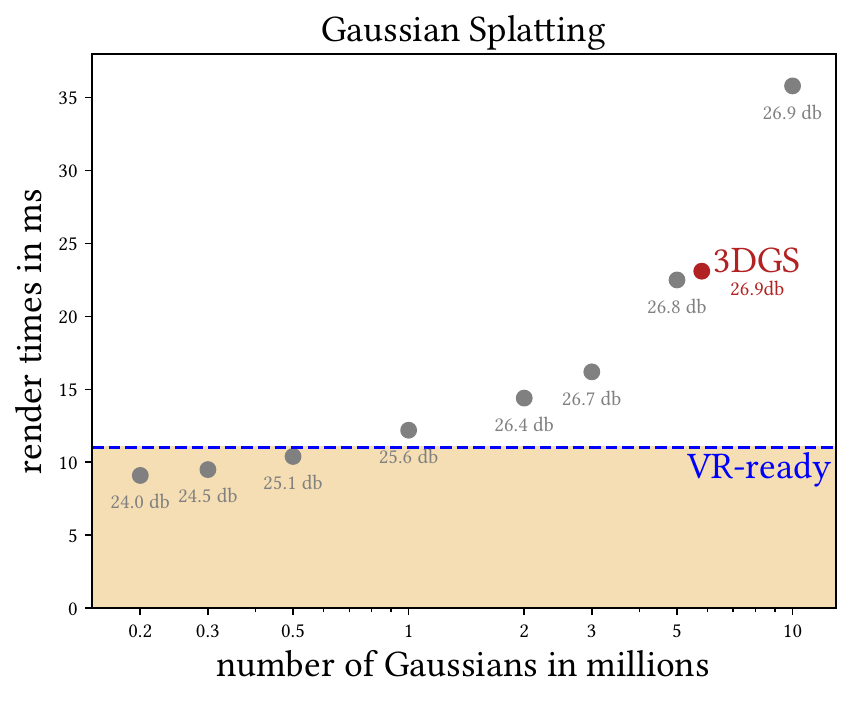}
    \small(a) GS: number of Gaussians compared to render time in full VR resolution, with PSNR in db
    \end{subfigure}\hfill%
    \begin{subfigure}[]{.48\linewidth}
    \centering
    \includegraphics[width=.99\linewidth]{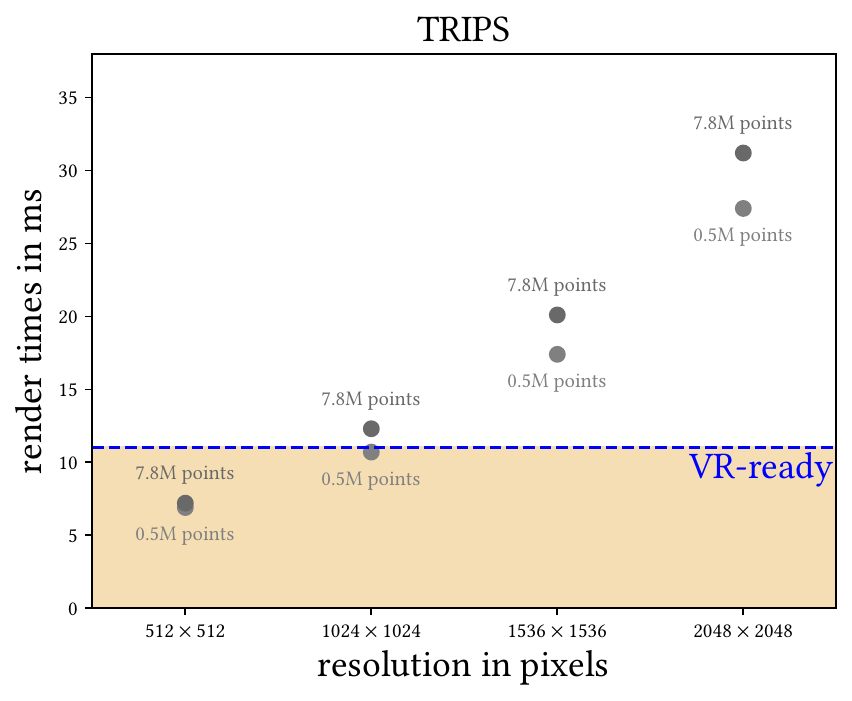}
    \small(b) TRIPS: resolution compared to render time, with number of primitives
    \end{subfigure}%
    \caption{Rendering efficiency of 3D Gaussian Splatting~\cite{kerbl3Dgaussians} and TRIPS~\cite{franke2024trips}.
    Left: Gaussian Splatting (GS) performance decreases drastically with increased number of primitives. To fit within VR limits, Gaussian counts need to be kept low, impacting quality.
    Right: In contrast, TRIPS performance scales mainly with resolution, allowing efficient rendering of high quality small crops.
    }
    \label{fig:motivation_plot}
\end{figure}

Recently, 3D Gaussian Splatting (3DGS)~\cite{kerbl3Dgaussians} emerged as a volumetric scene representation for fast and realistic novel view synthesis.
Its volumetric nature inherently produces smooth renderings, and it lacks aliasing due to alpha blending the primitives as well as low-pass filtering before rasterization.
However, to achieve pleasing results with crisp details, the required number of Gaussian primitives pushes the framerates below acceptable levels for VR, as shown in Figs.~\ref{fig:teaser} and \ref{fig:motivation_plot} (a).
On the other hand, neural point rendering-based methods, like TRIPS~\cite{franke2024trips}, promise sharper results for the same time budgets.
They use neural networks as sophisticated image filters that interpret colors from rasterized neural point colors (often called descriptors) and jointly fill holes originating from projecting small-sized point splats.
Both it and 3DGS use a Lagrangian scene representation, but model the splat sizes in different ways: 3DGS models splat size in Euclidean space via variance parameters, while the size of neural points is interpreted in pixel space using a neural network. 

This causes the rendering performance to scale differently.
3DGS requires tiling and sorting all primitives; as such, the rendering speed of 3DGS is reduced with an increasing number of primitives.
In contrast, neural point rendering methods rasterize millions of primitives fast but are highly sensitive to resolution increases due to the convolutional neural network~\textit{(CNN)} employed for screen-space filtering.
Fig.~\ref{fig:motivation_plot} illustrates this behavior.
Unfortunately, both methods do not meet VR framerate requirements when used with full resolution and their default number of primitives.

\begin{figure}
    \centering
    \includegraphics[width=1\linewidth]{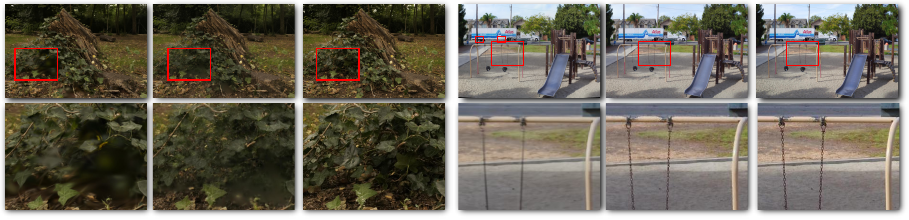}
    \begin{tabular}{@{}>{\centering\arraybackslash}p{0.15\linewidth}@{}
                >{\centering\arraybackslash}p{0.16\linewidth}@{} 
                >{\centering\arraybackslash}p{0.18\linewidth}@{} 
                >{\centering\arraybackslash}p{0.16\linewidth}@{} 
                >{\centering\arraybackslash}p{0.16\linewidth}@{} 
                >{\centering\arraybackslash}p{0.16\linewidth}@{}}
    \small 3DGS & \small TRIPS & \small Ground Truth & \small 3DGS & \small TRIPS & \small Ground Truth 
    \end{tabular}
    \caption{Equal render time comparison for components of our method. Fovea-sized crops highlighted.
    For fine details, TRIPS often shows crisper results.
    }
    \label{fig:trips_vs_gs}
\end{figure}

In this juxtaposition, we identify a sweet spot matching foveated rendering: 
For the fovea, neural point rendering is able to cheaply render high-detailed, crisp small image crops.
In particular, TRIPS has shown to be able to render fine details accurately (as seen in Fig.~\ref{fig:trips_vs_gs}), which is advantageous for perceived quality given the magnification effects of current VR headset lenses.
Regarding the periphery, Gaussian Splatting even with reduced number of primitives perfectly harmonizes with the demands of peripheral vision for high coherence and low flickering, as it renders a smooth but low detailed image which is temporally stable.

Instead of a purely 3DGS-based method, which reduces the number of primitives in the periphery, we opt for such a hybrid approach that marries the individual advantages of the chosen methods regarding fovea and periphery reconstruction. 
In this paper, we thus introduce a hybrid system, which also allows us to further accelerate the system through synergies.

Given a VR headset with integrated eye tracking, our method first renders a low-detailed Gaussian rendering for both eyes for the full field of view.
Hereby, we find that pixel-correct sorting~\cite{radl2024stopthepop} proves vital for rendering smoothly.
Using a occlusion estimation from this rendering together with querying the eye tracker, we cull the neural point cloud and rasterize it into an image pyramid for the foveal regions pixel footprint.
This pyramid, together with the cropped Gaussian rendering, is then resolved with a small CNN, allowing the network to focus on adding fine detail in the foveal area only.
The results of both renderings are then combined and sent to the VR headset.

Our novel Gaussian and neural point representation is optimized from captured input images end-to-end, allowing virtual walk-throughs through real-world environments.
We contribute:
\begin{itemize}
    \item A high-fidelity, 90 Hz capable technique using foveated radiance fields rendering for first-person VR rendering.
    \item The combination of neural point rendering and Gaussian rendering techniques for crisp foveal and smooth peripheral rendering.
    \item An evaluation of our foveated rendering method, including a user study showing its effectiveness.
\end{itemize}

\section{Related Works}

\subsection{Foveated Rendering}
Exploiting the limitations of the human visual system (HVS) is a well established approach to relax rendering constraints and reduce computational demands~\cite{wang2023foveated,weier2017perception}. 

\paragraph{The Human Visual System.}
A key characteristic of human vision is the decrease in visual acuity in our peripheral vision, which foveated rendering techniques leverage.
Weier et al.~\shortcite{weier2017perception} provide a comprehensive overview of the HVS and its exploitable limitations.

For designing rendering algorithms, the retina's physiology is crucial. 
The retina contains cones and rods that respond to light stimuli.
The cones, mainly concentrated in the fovea (less than $10^{\circ}$  of the visual field), are responsible for sharp, color vision~\cite{goldstein2016sensation}. 
Rods, on the contrary, are motion sensitive and less color sensitive, dominating the peripheral retina with the highest concentration around $17^{\circ}$ from the gaze direction~\cite{curcio1990human}. This distribution allows the human visual system to be modeled with a foveal region (\textit{fovea}) for sharp vision and a peripheral region (\textit{periphery}) sensitive primarily to brightness and motion.
Exploiting this model to accelerate the rendering is called \textit{foveated rendering}. 
Sometimes, it is assumed that the users gaze is always centered on the image, then called \textit{fixed foveated rendering}.
However, this assumption has failure cases and usually requires the foveal region to extend much larger than with eye-tracked methods.
In this work, we employ eye tracking, as this allows us to keep the foveal region small and achieve better performance.

\paragraph{Foveated Synthetic Rendering.}
The concept of foveated rendering has been extensively investigated in the context of synthetic, artist-created scene rendering~\cite{wang2023foveated, jabbireddy2022foveated}.
Guenter et al.~\shortcite{guenter2012foveated} proposed generating three images per frame with full resolution in the fovea and progressively lower resolution in the periphery, blending them using bilinear upsampling. 
Nevertheless, this requires robust anti-aliasing to mitigate flickering, which can be reduced by decoupling visibility detection from shading~\cite{patney2016towards}.

Adaptive sampling in foveated ray tracing~\cite{koskela2016foveated} also reduces the computational load by focusing more samples on the fovea and fewer on the periphery~\cite{friston2019perceptual,stengel2016adaptive,weier2016foveated,sun2017perceptually}. 
Stengel et al.~\shortcite{stengel2016adaptive} use saliency maps to allocate samples effectively, while Weier et al.~\shortcite{weier2016foveated} employ a reprojection scheme to accelerate ray tracing by resampling only visually important areas in the periphery, an idea also transferred to rasterization~\cite{franke2021time}.
Other techniques include log-polar mappings for cost reduction~\cite{meng2018kernel,meng20203d}, contrast-aware foveation for ray tracing and rasterization~\cite{tursun2019luminance} and deep foveated reconstruction trained with video~\cite{kaplanyan2019deepfovea} or neural super resolution techniques~\cite{ye2024neural}.

Further supporting research has explored methods to reduce perceptual artifacts. 
For example, adding depth of field~\cite{weier2018foveated}, developing perceptual image metrics~\cite{swafford2016user,mantiuk2021fovvideovdp,mantiuk2022stelacsf}, exploring saliency~\cite{sitzmann2018saliency}, exploiting eye domination~\cite{meng2020eye}, hardware solutions~\cite{kim2019foveated} and examining peripheral acuity~\cite{hoffman2018limits} have all proofed promising. 
Eye-tracking latency requirements~\cite{albert2017latency} suggest that a maximum latency of 50-70 ms is acceptable to avoid noticing artifacts, particularly during rapid eye movements (saccades). 

\subsection{Radiance Field Rendering}
Radiance fields are a common set of representations for novel view synthesis (NVS), by which new views are created from angles previously unseen.
\paragraph{Traditional Methods.}
Traditionally, NVS relied on light fields~\cite{gortler1996lumigraph}, but image-based rendering, often by warping source views onto geometric proxies~\cite{debevec1998efficient} has since emerged as a popular alternative~\cite{shum2000review}.
Although this rendering can be performed fast and efficiently, quality is highly dependent on the quality of the proxy geometry, although subsequent methods have sought to mitigate these issues~\cite{eisemann2008floating, chaurasia2013depth}.
Full 3D reconstructions, which gained traction with the advent of \textit{Structure-from-Motion (SfM)}~\cite{snavely2006photo,schoenberger2016sfm} and \textit{Multi-view Stereo (MVS)}~\cite{seitz2006comparison,schoenberger2016MVS,goesele2007multi}, also suffer from similar issues. 
Although requiring significantly more render time, neural rendering extensions~\cite{Tewari2022NeuRendSTAR} employing learned blending operators~\cite{hedman2018deep} and deep learned textures~\cite{thies2019deferred} in NVS strongly increase quality.

\paragraph{Neural Radiance Fields.}
More recently, implicit 3D scene representations using volumetric fields have become popular, enabling novel view synthesis through volume rendering without requiring proxy geometry. 
The \textit{Neural Radiance Field (NeRF)} introduced by Mildenhall et al.~\shortcite{mildenhall2020nerf} has shown remarkable results by encoding entire 3D scenes into a coordinate-based MLP. 
Subsequent research has addressed challenges related to input view distributions~\cite{chibane2021stereo,yu2021pixelnerf,kopanas2023improving,wu2024reconfusion}, scaling~\cite{turki2022mega,tancik2022block,mi2023switchnerf} and computation times~\cite{barron2021mipnerf, neff2021donerf, chen2021MVSnerf, chibane2021stereo, mueller2022instant, tancik2021learned, barron2023ICCV,barron2022mipnerf360}. 
Effective strategies include scene space discretization through voxel grids~\cite{Fridovich2022}, octrees~\cite{ruckert2022neat,yu2021plenoctrees}, tensor decomposition~\cite{chen2022tensorf}, and faster model inference through neurally textured triangle meshes~\cite{chen2023mobilenerf} or mesh baking~\cite{yariv2023bakedsdf,reiser2024binary,reiser2023merf,duckworth2023smerf}. 
Notably, M\"uller et al.~\shortcite{mueller2022instant} achieved significant improvements in training and rendering speed using a hash grid-based space partitioning scheme, allowing fast training and framerates of up to 10 fps.

\paragraph{Point-based Radiance Fields.}
Compared to NeRFs, point-based radiance fields require an explicit point cloud proxy captured from LiDAR~\cite{liao2022kitti}, RGB-D cameras with depth fusion~\cite{dai2017bundlefusion,whelan2016elasticfusion, keller2013realtime} or estimated through SfM or MVS.
Point clouds represent an unstructured set of spatial samples with varying distances between neighbors, closely reflecting the original captured data. 
Rendering these point clouds is highly efficient~\cite{schutz2021rendering, schutz2022software, schutz2019real}, and when augmented with neural descriptors~\cite{ruckert2022adop, aliev2020npbg, rakhimov2022npbgpp, franke2023vet, hahlbohm2023plenopticpoints, franke2024trips, hahlbohm2024inpc, kappel2024d} or as a proxy for optimized attributes~\cite{kopanas2021perviewopt, kopanas2022catacaustics}, they can produce high-quality images using differentiable point renderers~\cite{wiles2020synsin, yifan2019differentiable} or neural ray-based renderers~\cite{xu2022point, ost2022neural, abou2024particlenerf}. 

A significant issue in point rendering is filling the gaps in the rendered images, which has led to the development of two main approaches: \textit{world-space splatting} and \textit{screen-space hole filling}~\cite{kobbelt2004survey}. 
In world-space splatting, points are rendered as oriented discs or 3D structures, known as "splats" or "surfels", with radii precomputed based on point cloud density.
To minimize artifacts, these splats are often combined using Gaussian alpha masks and normalizing blend functions~\cite{alexa2004point, pfister2000surfels, zwicker2001surface}. 
Recent innovations in this area include optimizing splat sizes~\cite{kerbl3Dgaussians, zhang2022differentiable} and employing neural networks to improve rendering quality~\cite{Yang_2020_CVPR}.
The primary radiance field methology using splats is \textit{3D Gaussian Splatting (3DGS)}~\cite{kerbl3Dgaussians}, which sparked a plethora of subsequent work~\cite{chen2024survey,wu2024recent}, including algorithmic extensions~\cite{yu2024mip,Huang2DGS2024,kheradmand20243d,radl2024stopthepop,hahlbohm2025htgs,mallick2024taming}, large-scale variants~\cite{ren2024octreegs, kerbl2024hierarchical}, robotics~\cite{meyer2024pegasus}, and dynamic content~\cite{luiten2023dynamic,yang2023deformable,huang2023sc}.
These methods are able to render high-quality content at high framerates.

Screen-space hole filling commonly uses small splats at most a few pixels large and fills the resulting sparse image in screen space~\cite{aliev2020npbg}. 
Therefore, a deep filter is commonly used, in the form of a U-Net~\cite{aliev2020npbg}, possibly adapted with Fourier convolution blocks~\cite{hahlbohm2024inpc,zhu2024rpbg} or a decoder-only CNN~\cite{franke2024trips}.
Extensions in this domain include reflection models~\cite{kopanas2022catacaustics}, scene optimization~\cite{franke2023vet,hahlbohm2023plenopticpoints, hahlbohm2024inpc, zhu2024rpbg}, reduced training regimes~\cite{harrerfranke2023inovis, rakhimov2022npbgpp}, modeling photometric properties and lens distortion~\cite{ruckert2022adop}, and linear splatting formulations~\cite{franke2024trips,hahlbohm2024inpc}.

\paragraph{VR and Foveated Radiance Field Rendering.}
Radiance field rendering methods for VR are sparse, especially due to rendering budget constraints exceeding common methods.

In the domain of NeRFs, FoV-NeRF~\cite{deng2022fovnerf} introduces an egocentric NeRF variant with adapted sampling schemes for foveated rendering using multiple MLPs.
Shi et al.~\shortcite{shi2024scene} extend this with a scene-aware formulation, whereby they improve the expressiveness for detailed areas.
Wang et al.~\shortcite{wang2024vprf} further augment this field with a voxel-based representation including a visual sensitivity model.
These three approaches inherit the challenging inference performance characteristics of NeRFs, as framerates stays below 90 Hz.
Similarly, hash grid NeRF-based super-resolution techniques~\cite{li2022immersive} and variable rate shading approaches~\cite{rolff2023vrs,rolff2023interactive} achieve impressive results, but still fall short of the high framerate and resolution requirements of VR rendering.
Consequently, Xu et al.~\shortcite{xu2023vr} introduce a multi-gpu rendering scheme, allowing 36 fps when using three A40 GPUs.
Additionally, they rely on a space-carving scheme to lower ray-marching costs.
Following, Turki et al.~\shortcite{turki2024hybridnerf} extend this with a hybrid representation reducing rendering costs for surfaces and only using full ray-marching for challenging regions.

Point-based methods also struggle with VR capabilities. Neural point-based methods such as ADOP~\cite{ruckert2022adop} enable VR rendering, however only for low resolution output.
Likewise, the recently published official SIBR~\cite{sibr2020}-based VR implementation, as well as Gaussian Splatting extensions for Unity and Unreal, fail to render high-resolution details due to the amount of Gaussians required. 
Although some methods focus on VR applications~\cite{jiang2024vr}, they do not prioritize rendering efficiency.

In our method, we present the first method focusing on achieving high fidelity and fast VR rendering of radiance fields on consumer-grade hardware.

\section{Method}
\label{sec:method}
As input, we use a set of real-world images captured using, e.g., a smartphone or DSLR camera.
For our method, we use SfM/MVS~\cite{schoenberger2016sfm,schoenberger2016MVS} to obtain camera intrinsics, poses, and point clouds. 
We first present and evaluate the method core (Sec.~\ref{sec:prototype}), motivating the refinements with more sophisticated peripheral rendering (Sec.~\ref{sec:gaussian_periphery}), accelerated foveal rendering (Sec.~\ref{sec:fovea_neural_points}), and a fovea and edge-based combination function (Sec.~\ref{sec:combination}).
Our pipeline can be seen in Fig.~\ref{fig:pipeline} and is optimized end-to-end from the input images, camera parameters, and point clouds (Sec.~\ref{sec:losses}).

\subsection{Pilot Study}
\label{sec:prototype}

\newcommand{\E}{\textit{E}}
\newcommand{\En}[1]{\textit{E#1}}

The core idea is a deftly blend of Gaussian splats and neural points for foveated rendering. Similar to Patney et al.~\shortcite{patney2016towards} using a \textit{perceptual sandbox}, we first implemented a simplistic form of this combination and evaluated it with a pilot study, in order to evaluate the potential of the idea and to identify problems that require further refinement. 
First, we present and discuss the findings of this first naive baseline.
Subsequently, we continue with the description of our final system, including a number of improvements, which we developed based on the study findings.

\paragraph{Naive Method.}
To render a view, we render the point cloud in a foveated fashion, using fixation point information obtained from the VR headset's eye tracker.

For the periphery, we employ 3D Gaussian Splatting~\cite{kerbl3Dgaussians}.
Each splat has an opacity $\alpha$ and a color $C$. 
Their 3D orientation and scaling are parameterized per point via a covariance matrix $\Sigma$.
These primitives are projected and blended to screen, resulting in a full-resolution image $P$.
For the study, the splats were initialized from SfM and densified carefully during training to ensure low rendering cost, resulting in 0.4 million primitives.

For the foveal region, we employ TRIPS~\cite{franke2024trips} initialized with a dense point cloud estimated by MVS.
Each point has contribution size~$s$ and a compressed neural feature~$\tau$.
The features essentially represent colors; however, their higher dimensionality allows encoding additional information.
The points are projected and blended into a multi-resolution image pyramid, based on the projected size $s$~\cite{franke2024trips} and then combined by a small CNN, resulting in a fovea-sized image $F$.
For the pilot study, $P$ and $F$ were combined using a smooth blending function and the fovea size was set to $17^{\circ}$.

The parameters of Gaussians, neural points and CNN are optimized end-to-end with gradient descent from the input images with their respective proposed optimization functions.

\paragraph{Pilot Study.}
We asked four computer graphics experts (aged 23-28, 1 female, 3 male) with VR experience for their observations and rankings on a VR-ready 3DGS version with 0.5M primitives (denoted as \textit{VRGS}) and contrasted it with our naive baseline on the popular \textsc{Garden} scene from the MipNeRF-360 dataset~\cite{barron2022mipnerf360}.

\begin{figure}
    \centering
    \includegraphics[width=\linewidth]{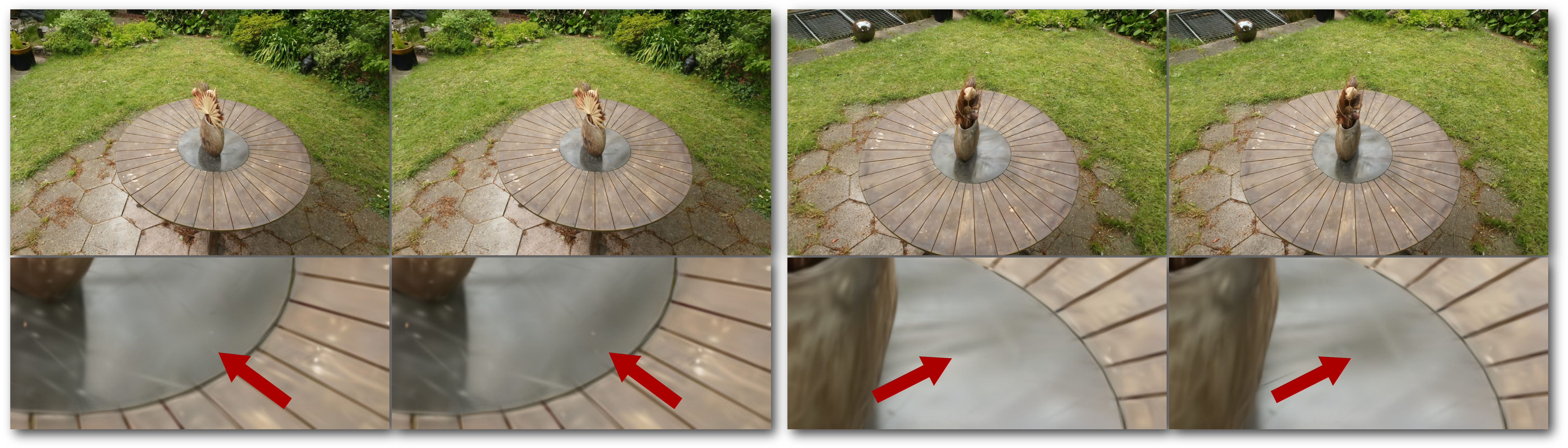}
    \caption{Gaussians in 3DGS "popping" in and out during small camera rotations, resulting in black spots suddenly appearing and disappearing. Popping is especially noticeable in VR, as revealed in our pilot study.}
    \label{fig:popping}
\end{figure}

The participants were asked to state their most important visual observations. 
Consistent for both stimuli, participants noticed distracting "popping" artifacts, caused by inaccurate depth sorting in 3DGS, as noted by Radl and Steiner et al.~\shortcite{radl2024stopthepop}. 
An example is shown in Fig.~\ref{fig:popping}.
The effect is amplified by the magnification of the VR headset's lenses and the reduced number of primitives. 

Regarding VRGS, participants noted blurred splotches due to underreconstruction with the reduced number of Gaussians~\cite{ye2024absgs}.
More blurriness compared to the foveated method was also pointed out, which coincides with observations from prior works~\cite{franke2024trips,hahlbohm2024inpc}.

For our foveated method, the quality of the visual experience was rated slightly lower compared to VRGS.
Although some mentioned increased detail richness, the experience was overshadowed by artifacts. %
Several things were mentioned: mismatching color palettes as the VGG-loss used by Franke et al.~\shortcite{franke2024trips} is known to not force color fidelity~\cite{zuo2022snp}.
Furthermore, the suboptimal response of eye tracking due to delays and blinking was criticized.
Lastly, noise and temporal jittering was mentioned, also noted in related work~\cite{ruckert2022adop,aliev2020npbg}.

The study confirmed that the performance cuts of the VR-ready version of Gaussian Splatting lead to apparent artifacts due to underreconstruction.
Although the experience of the naive foveated implementation was lacking for users, it was rated at a similar quality level.
However, the experts' critique showed that combining peripheral and foveal regions from two different representations is non-trivial.

\begin{figure}
    \centering
    \includegraphics[width=0.99\linewidth]{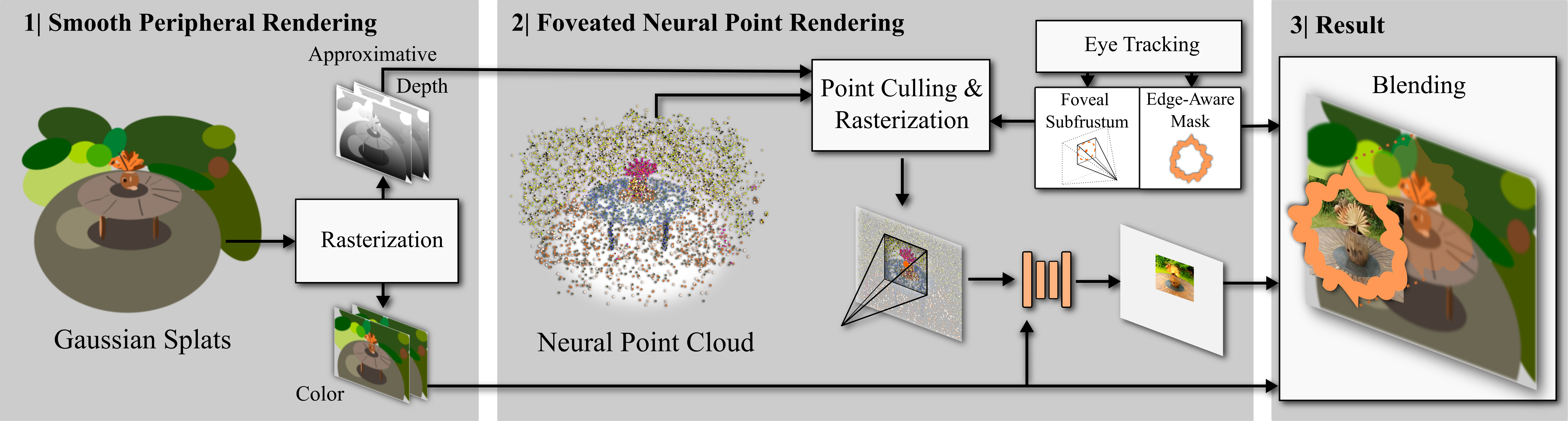}
    \caption{Our Pipeline. A smooth color image and approximate depth map are rendered from a limited set of 3D Gaussians with a temporally stable sorting active (see Section~\ref{sec:gaussian_periphery}). 
    Afterwards the eye tracking system is queried to construct a subfrustum via an adapted projection matrix covering only the foveal region.
    We project a separate neural point cloud with the adapted matrix.
    Points occluded by the denser Gaussian splats, are culled against the approximate depth maps and the result is processed by a small CNN (see Section~\ref{sec:fovea_neural_points}).
    Eventually, we blend the peripheral color output and the foveal region using an egde-aware mask (see Section~\ref{sec:combination}). 
}
    \label{fig:pipeline}
\end{figure}

Based on these findings, we identify three improvements: smoother Gaussian reconstruction for peripheral regions without popping artifacts (Sec.~\ref{sec:gaussian_periphery}), fast foveal reconstruction with well-integrated eye tracking (Sec.~\ref{sec:fovea_neural_points}), and a coherent mixture of the models including a matching color palette and subtle blending of high-frequency detail (Secs.~\ref{sec:combination} and~\ref{sec:losses}).
Our resulting pipeline can be seen in Fig.~\ref{fig:pipeline}.

\subsection{Peripheral Rendering}\label{sec:gaussian_periphery}

We identify three key properties which are required for our Gaussian-based peripheral reconstruction: (1) high rendering performance, (2) little-to-no popping artifacts, and (3) little underreconstruction.
Usually, these three are conflicting goals meaning an acceptable solution for all three needs to be designed carefully.

Regarding (2), we adapt the sorting scheme of Radl and Steiner et al.~\shortcite{radl2024stopthepop}, which eliminates popping by hierarchically sorting Gaussian contributions per pixel (instead of per primitive) at the cost of increased processing time per Gaussian.
Furthermore, by employing a more finely detailed Gaussian densification heuristic~\cite{kerbl2024hierarchical, ye2024absgs}, we reduce the severity of (3).

\begin{figure}
    \centering
    \includegraphics[width=0.8\linewidth]{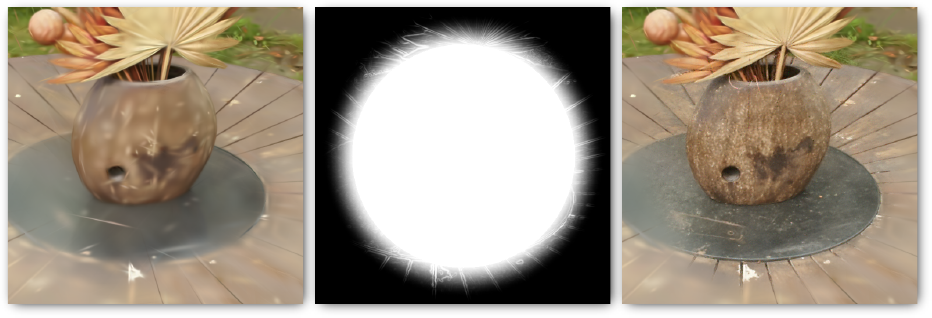}
    \begin{tabular}{*{3}{p{0.23\linewidth}<{\centering}}}
    \small (a) Gaussian rendering $P$ & \small(b) Combination blending map & \small(c) Our result combining $P$ and $F$
    \end{tabular}
    \caption{Our combination mask (b) in effect (zoomed in). It is used to adaptively merge Gaussian output with neural point rendering. }
    \label{fig:combine_fig}
\end{figure}

This Gaussian renderer is temporally stable also with regards to popping and without underreconstruction artifacts which are noticeable in the periphery.
An example rendering can be seen in Fig.~\ref{fig:combine_fig}~(a).
Regarding performance, we need to drastically reduce the number of Gaussians (as seen in Tab.~\ref{fig:motivation_plot}), which we achieve by limiting the densification during training.

Note that we do not mask out the foveal region in this step and instead generate a full image.
Furthermore, we generate an approximate depth buffer~\cite{radl2024stopthepop} by accumulating per-pixel depths $d$ for the number of all contributing Gaussian $n$ with
\begin{equation}
\label{eq:depth_gs}
    d = \sum_{i=1}^n d_i  \alpha_i  \prod_{m=1}^{i-1}(1-\alpha_m).
\end{equation}
While this depth buffer is not needed for the peripheral rendering, it is used to cull occluded points in the following foveal point rendering step.

\subsection{Foveal Rendering}\label{sec:fovea_neural_points}

In our context of foveated rendering, we improve TRIPS~\cite{franke2024trips} significantly by using information gained with the previous (full screen) Gaussian rendering results.

Firstly, we only render a subfrustum of the entire scene and discard points outside this frustum as early as possible.
This subfrustum is constructed around the fixation point, which is queried from the eye tracker after the periphery has been rendered.
Additionally, we use the approximate depth buffer rendered (Eq.~\ref{eq:depth_gs}) from the Gaussians to discard occluded points.
This occlusion culling reduces rendering times, accelerates convergence, and eliminates point render flickering, as this simplifies visibility and blending computations.

Furthermore, we inject the fovea crop of the less detailed Gaussian rendering into the CNN.
Thus, the network is only tasked with augmenting fine details, instead of balancing hole filling and crisp reconstruction as in TRIPS.
As a side product, this severely reduces training cost. 

These optimizations allow us to halve the number of filters for the largest-resolution CNN layer to further accelerate our pipeline.
Largest-resolution convolutions are originally the most expensive part of the TRIPS network evaluation, taking up more than 70\% of the network inference time.

\subsection{Combination}\label{sec:combination}

Our renderer outputs two images $P$ (peripheral) and $F$ (foveal).
It is important to carefully combine both images, as the visual acuity falloff of the human eye is gradual.
Therefore, we introduce a combination factor $c$ that includes the two terms $f_p$ (positional) and $f_e$ (edge-based) for each pixel (u,v):

\begin{equation}
    r_{norm} = \text{clamp}\left(\frac{1}{d_f} \sqrt{(u-e_x)^2 + (v-e_y)^2},0,1\right)
\end{equation}
\begin{equation}
    f_p =  \frac{r_{norm} - m}{1-m},
\end{equation}
with $e_{x,y}$ being the pixel position of the fixated point, $d_f$ the fovea radius in pixels and $m$ the start of the normalized blending interval between 0 and 1. 
We empirically use $m=0.75$.
We use $d_f$ to 256 pixels (see Sec.~\ref{sec:impl_detail}), which is larger than the fovea's 7\% eccentricity, however necessary due to inaccuracies of the eye-tracker.

The second factor is the edge factor $f_e$, based on the Sobel edge filter $S$ of the point image $F$:
\begin{equation}
    f_e = S(F,u,v).
\end{equation}
$f_e$ is inspired by Franke et al.~\shortcite{franke2021time}, who force color discontinuities in the transition between the fovea and the periphery to be gradual to avoid fast color changes.

Finally, the interpolation factor per pixel $c$ is computed as 
\begin{equation}
    c = 1 - S_2(f_p + \gamma * f_e), 
\end{equation}
using $\gamma = 0.2$ and with $S_2$ the smootherstep function $S_2(x) = 6x^5-15x^4+10x^3$ for a smooth transition and well defined derivations suited for end-to-end training and smooth blending.
The combined image is then computed for each pixel with $(1-c)p + cf$ with $p$ and $f$ the individual pixels of $P$ and $F$.

To account for variance in image exposures, we then follow ADOP~\cite{ruckert2022adop} and use a differentiable camera model to tone-map the resulting images to a displayable LDR space.

\subsection{Losses and Optimization}\label{sec:losses}

We optimize our pipeline end-to-end.
For every iteration, we randomly sample a fixation point and evaluate the pipeline to arrive at a foveated image.
We compare this image to the ground truth and optimize via gradient descent using our loss function $\mathcal{L}$:
\begin{equation}
    \mathcal{L} = (1-\lambda)\mathcal{L}_1 + \lambda\mathcal{L}_{D-SSIM} + \mu\mathcal{L}_{VGG}(F) + \beta \mathcal{R}
\end{equation}
The first two terms are identical to 3DGS~\cite{kerbl3Dgaussians} with $\lambda=0.2$.
We further augment this with the VGG19-loss~\cite{vgg19loss} for neural point rendering with $\mu=0.001$.
VGG-loss has proven to work well for neural point rendering; however, for Gaussian Splatting, it increases convergence time and causes artifacts with densification.
Therefore we only apply it to neural point renderings.
Lastly, we introduce a regularizer $\mathcal{R}$:
\begin{equation}
    \mathcal{R} = \lvert \overline F -  \overline{\mathcal{F}(P)} \rvert ,
\end{equation}
with the foveal cropping function $\mathcal{F}$.
This ensures that the mean of $F$ and the part of $P$ with pixels inside the fovea are similar.
This regularizer is vital to reduce the color shift, which optimizing with VGG loss otherwise induces.
We empirically found $\beta = 0.00001$ to work well.
Although a slight shift is still present, it isn't perceived by users.

We apply an opacity penalty to the opacities of both Gaussians and neural points~\cite{franke2023vet,radl2024stopthepop}, which causes the optimization to discard obsolete primitives.

\section{Implementation and Optimization Details}\label{sec:impl_detail}

We implemented our method in C++ and CUDA, using torch as the automatic gradient framework and followed the framework of Patas~\shortcite{patas23gscuda} for a Gaussian Splatting implementation and Radl and Steiner et al.'s sorting algorithm~\cite{radl2024stopthepop}.

\paragraph{Training.}
For training, we optimize Gaussians for 30000 iterations (as in 3DGS~\cite{kerbl3Dgaussians}) and neural points for 30000 additional iterations with fixed Gaussian parameters.
For the first 2000 iterations, both systems are trained independently to avoid the faster converging Gaussians to interfere with neural point optimization.
Afterwards, gradients from the foveal rendering are also propagated back to Gaussian parameters except for VGG-loss computations.
Sampling a random eye position every iteration maps well to cropped training usually employed with CNNs.
After 45000 iterations, we start regularizing the training with $\mathcal{R}$.

For densification, we lower the thresholds compared to 3DGS and use $0.0002$ as densification gradient threshold, $200$ as densification interval, $0.005$ as opacity pruning threshold, and $0.999$ as opacity decay every 50 epochs~\cite{radl2024stopthepop}.
For the scenes we tested, this results in 0.3M to 0.5M Gaussians.
For neural points, we impose an opacity gradient penalty of $10^{-8}$ every update step~\cite{franke2023vet}, usually resulting in about 5M points.

We uniformly sample fixation points over the image, which is similar to TRIPS using a cropped training regime.
The increased iterations hereby also ensure all parts of the scene to receive coverage during training.

\paragraph{Inference.}
For inference, we adjusted several aspects to VR rendering.
We target the HTC Vive Pro Eye, having an eye-tracker capable of updating rates of 120 Hz.
We bake all rendering attributes directly into the point cloud, such as opacities, features, and covariances to minimize runtime costs.
Furthermore, we delay eye tracking updates until after rendering $P$, drastically reducing lag.
Also, compute and evaluate all renderings in half-precision and evaluate the network with images for both eyes batched together using cuDNN for acceleration.

For neural point rendering, rendering only the foveal regions allows for algorithmic optimizations with respect to performance:
TRIPS originally renders up to eight progressively lower layers individually, first by counting the pixel contributors, then scanning and allocating a buffer for splatting.
These steps introduces overhead, which we reduce by combining all layer computations, speeding up the computations by about 2 ms. 
This requires about four times the GPU memory of the original implementation; however, this is negligible as rendering is limited to the foveal region.

\paragraph{Fovea Size in Pixels.} 
The effective fovea size in pixels required for our formulation necessitates the estimation of pixels per degree of the field of view.
This is complicated by the use of distortion lenses~\cite{kreylos2018displayresolution} and is rarely accurately reported by headset manufacturers.
We follow Kreylos' upper limit estimate of 15.7 pixels per degree~\cite{kreylos2019vivepro} for the headsets display system, which worked well in our tests.
Related works use fovea sizes between $7.5^{\circ}$ and $17^{\circ}$~\cite{patney2016towards,franke2021time,weier2016foveated}.
Using the latter as the upper limit and the driver-initiated resolution scale of $1.4\times$, we arrive at a fovea size of around 400 pixels, which we round to 512 pixels (thus $d_f=256$) for implementation reasons and to combat eye-tracking inaccuracies.

\section{Evaluation}

We evaluate our method on real-world captured scenes; we use the popular MipNeRF-360 dataset~\cite{barron2022mipnerf360} as well as the \textsc{intermediate} set from Tanks\&Temples~\cite{Knapitsch2017}.

\subsection{Rendering Performance}

We compare our methods rendering performance with the established NeRF-based approaches FoV-Nerf~\cite{deng2022fovnerf}, Scene-aware Foveated Neural Radiance Fields~\cite{shi2024scene} (SA Fov-Nerf), VR-NeRF~\cite{xu2023vr} and HybridNeRF~\cite{turki2024hybridnerf}.
Furthermore, we include the differentiable point rendering techniques TRIPS~\cite{franke2024trips} and 3DGS~\cite{kerbl3Dgaussians} on which our method is built.

\begin{table}[]
\caption{\label{tab:rendering_compare} Comparison of reported rendering times of related VR-capable radiance field methods on a single RTX 4090. 
Timing $\dagger$ from Shi et al.~\shortcite{shi2024scene} and $\ddagger$ from Turki et al.~\shortcite{turki2024hybridnerf}.}
\tiny
\begin{tabular}{l|ccccccc}
                   & Fov-NeRF            & SA Fov-NeRF        & VR-NeRF            & HybridNeRF        & TRIPS              & 3DGS                 & Ours               \\
                   & \cite{deng2022fovnerf} & \cite{shi2024scene} & \cite{xu2023vr} & \cite{turki2024hybridnerf}& \cite{franke2024trips} & \cite{kerbl3Dgaussians} & \\\hline
Resolution         & $1440 \times 1700$ & $1440 \times 1700$ & $2064 \times 2096$ & $2064 \times 2096$ & $2016 \times 2240$ & $2016 \times 2240$ & $2016 \times 2240$ \\
Render times & 23.5 ms $^\dagger$              & 16.1 ms $^\dagger$               & 165.3 ms $^\ddagger$              & 21.8 ms $^\ddagger$              & 33.1 ms               & 23.1 ms               & 10.9 ms              
\end{tabular}
\end{table}

In Tab.~\ref{tab:rendering_compare} we present the results of this overview, measured on a single RTX 4090, with metrics reported by related works~\cite{shi2024scene,turki2024hybridnerf}.
Our performance calculations were performed by rendering in full VR resolution, with the drive- recommended $2016\times2240$ pixels per eye.
Our method is the only that surpasses the VR framerate limit of 11.1ms per frame, with speedups of more than $1.45\times$ compared to the fastest related approach ``SA Fov-Nerf", who report their timing only for resolutions of 2.44 megapixel compared to our 4.52 megapixel.

For further evaluation, we also compare to a reduced Gaussian Splatting variant with VR-ready performance using 0.5M primitives (see Fig.~\ref{fig:motivation_plot}).
We call this variant \textit{VRGS}, which renders slightly faster than our method (as seen in Tab.~\ref{tab:quantitative}).

For training speed, we improve drastically on TRIPS (see Tab.~\ref{tab:quantitative}, right), however, both Gaussian Splatting variants converge faster as no neural network optimization is required.

\subsection{Quantitative Evaluation}

To quantitatively evaluate our method, we compare our method with the mentioned VRGS, our interpretation of a VR-ready version of Gaussian Splatting.
Furthermore, we compare against 3DGS~\cite{kerbl3Dgaussians} and TRIPS~\cite{franke2024trips} as qualitative baselines and closest to our method even though they are not fast enough for VR.

\begin{table*}[]
\centering
\caption{Quantitative metrics. JOD (Just-Objectionable-Difference)~\cite{mantiuk2021fovvideovdp} is computed with \textit{FovVideoVDP} for an HTC Vive Pro in foveated mode for the full images. LPIPS, PSNR, and SSIM metrics are evaluated for the foveal regions only.}
\tiny
\begin{tabular}{ll|c|c|cccc|cccc|c}
&&&& \multicolumn{4}{c|}{\textit{MipNeRF-360}}& \multicolumn{4}{c}{\textit{Tanks\&Temples}} \\
 Target System&  Method  & No Popping & Performance & JOD$\uparrow$ & LPIPS$\downarrow$ & PSNR$\uparrow$& SSIM$\uparrow$& JOD$\uparrow$& LPIPS$\downarrow$ & PSNR$\uparrow$& SSIM$\uparrow$ & Training \\\hline
\multirow{2}{*}{ \rotatebox{0}{Desktop}} & 3DGS & $\times$ & 23.1\,ms &  7.90& 0.247 &  \textbf{27.01} &  \textbf{0.804}  &\textbf{8.24}& 0.198 &  \textbf{27.15} &  \textbf{0.874} & 0.5h\\

&TRIPS & $\checkmark$  &   33.1\,ms   &   \textbf{8.06}&  \textbf{0.222} & 25.19 & 0.761 &      8.15&   \textbf{0.184}& 24.70&0.835 &2.5h\\\hline

\multirow{2}{*}{ \rotatebox{0}{VR}}&VRGS & $\times$  &   10.4\,ms &  7.74 &  {0.286} &  \textit{26.45} &  \textit{0.765} & 8.11 &  {0.211} &  \textit{26.73} &  \textit{0.854} & 0.4h\\
&Ours &  $\checkmark$ &   10.9\,ms  & \textit{8.02}&  \textit{0.237} &  {26.03} &  {0.755}& \textit{8.16} & \textit{0.196} &  {25.97} &  {0.815} & 1.4h\\
\end{tabular}
\label{tab:quantitative}
\end{table*}

We compare using the common quality metrics $\text{LPIPS}_\text{VGG}$~\cite{zhang2018lpips}, PSNR and SSIM.
For these metrics, we only evaluate the foveal region and omit the periphery. 

The results are seen in Tab.~\ref{tab:quantitative}, averaged per dataset. 
Our experiments show that our method is preferable in the LPIPS metric, which is more perceptually close to the human visual system~\cite{zhang2018lpips,zuo2022snp}, while being close in the other metrics.
Importantly, our method is not subject to popping.

For computing JOD~(Just-Objectionable-Difference), we use FovVideoVDP~\cite{mantiuk2021fovvideovdp}, which is a perceptional metric that includes terms for contrast sensitivity and peripheral vision.
This method is designed and capable of metrically evaluating foveated rendering and scoring it in JOD between 0 and 10, the higher the better.
We use it with the preset for the HTC Vive Pro (having the same display system we use) in the setup: \textit{"FovVideoVDP v1.2.0, 13.15 [pix/deg], Lpeak=133.3, Lblack=0.1 [cd/m2], foveated"}.
The metric is computed including peripheral regions.
As also seen in Tab.~\ref{tab:quantitative}, our method consistently outperforms VRGS in this metric.

For VR renderings of our method, see the supplemental video and the project page.

\subsection{User Study}
To further evaluate our method against the VR-ready Gaussian Splatting VRGS, we conducted a user study.
In our study design, we follow Deng et al.'s foveated radiance field user study design and evaluation~\cite{deng2022fovnerf}.
The experiments were approved by an institutional ethical review board.

\begin{figure}
    \centering
    \includegraphics[width=\linewidth]{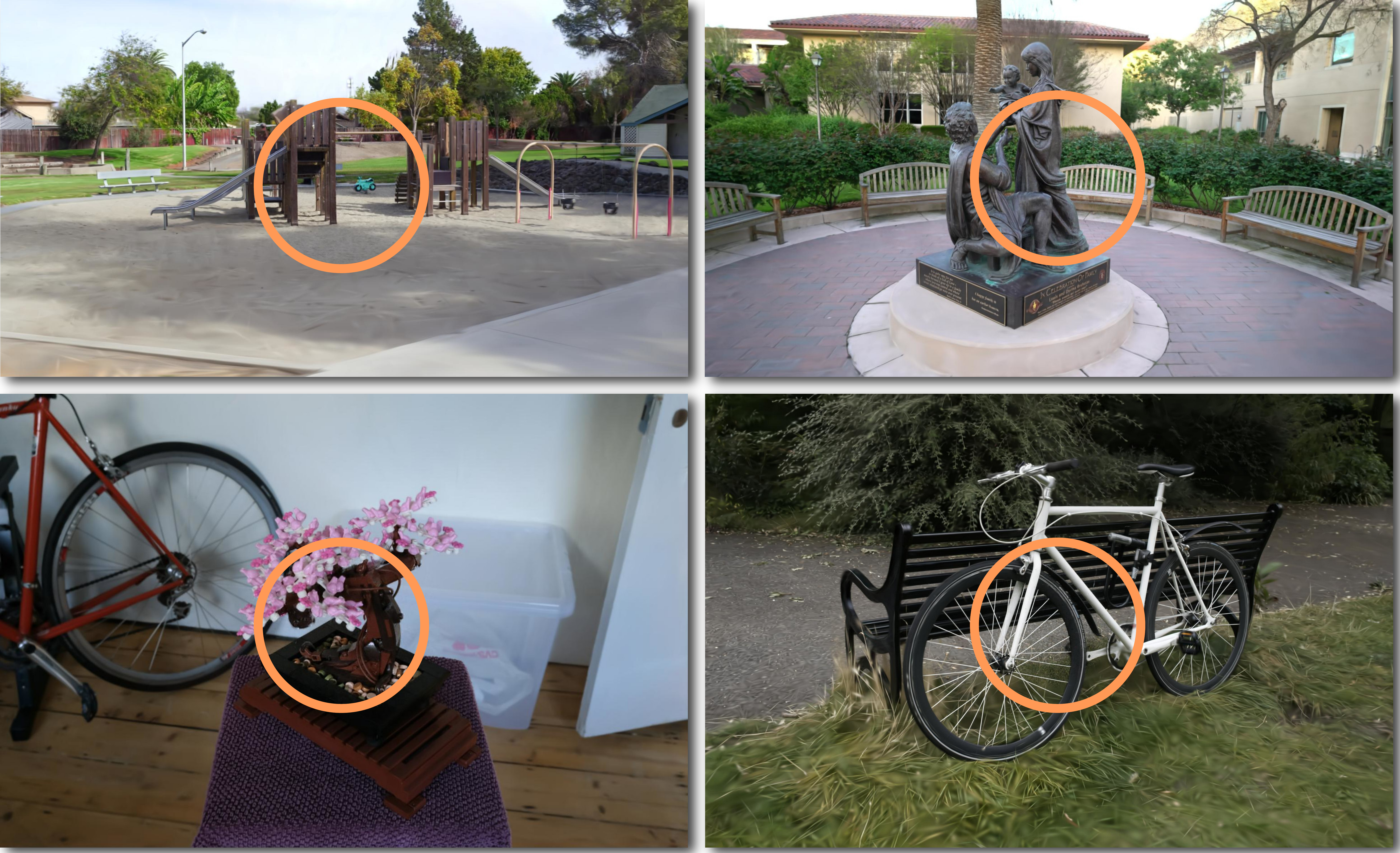}
    \caption{Example renderings of the \textsc{playground},  \textsc{family} (upper row), \textsc{bonsai}, and \textsc{bicycle} (lower row) scenes used in the user study. }
    \label{fig:user-study-examples}
\end{figure}

\paragraph{Stimuli.}
Each stimulus set comprised VR-ready stereo rendering using both our method and VRGS. The resolution of each image was $2016\times2240$ pixels per eye, and seated free exploration was possible.
We used scenes \textsc{garden}, \textsc{bicycle} and \textsc{bonsai} from the MipNeRF-360 dataset~\cite{barron2022mipnerf360} as well as scenes \textsc{family} and \textsc{playground}  from the Tanks\&Temples dataset~\cite{Knapitsch2017}.
Example renderings are presented in Fig.~\ref{fig:user-study-examples} and in the teaser (Fig.~\ref{fig:teaser}).

\paragraph{Setup.}
Each participant was seated and equipped with an eye-tracked HTC Vive Pro Eye headset to view stimuli throughout the experiment.
The eye tracker was calibrated for each participant with the default calibration kit.
The users had a keyboard in front of them to switch rendering modes at will as well as to confirm the choices.
Participants could switch the rendering mode by pressing the spacebar and confirm their choice with the enter button.
In total, twelve users (aged 22 to 30, 4 female, 8 male) participated in the experiment.

\paragraph{Task.}
The task followed a two alternative forced choice (2AFC) design.
Each trial presented a pair of stimuli generated by our method and VRGS, with the order of viewpoints and the first mode displayed being randomized. 
A mandatory $0.5$ second break~\cite{guenter2012foveated}, displaying a gray screen, was introduced between conditions to reset visual perception, with the same applied between different scenes.

Each participant completed 20 trials throughout the experiment. 
Before the experiment, the participants were briefed and the device was calibrated. 
During the study, participants were allowed to take as long as needed to complete each test.
The participants were unaware of the experimental hypothesis.

\begin{figure}
    \centering
    \includegraphics[width=.6\linewidth]{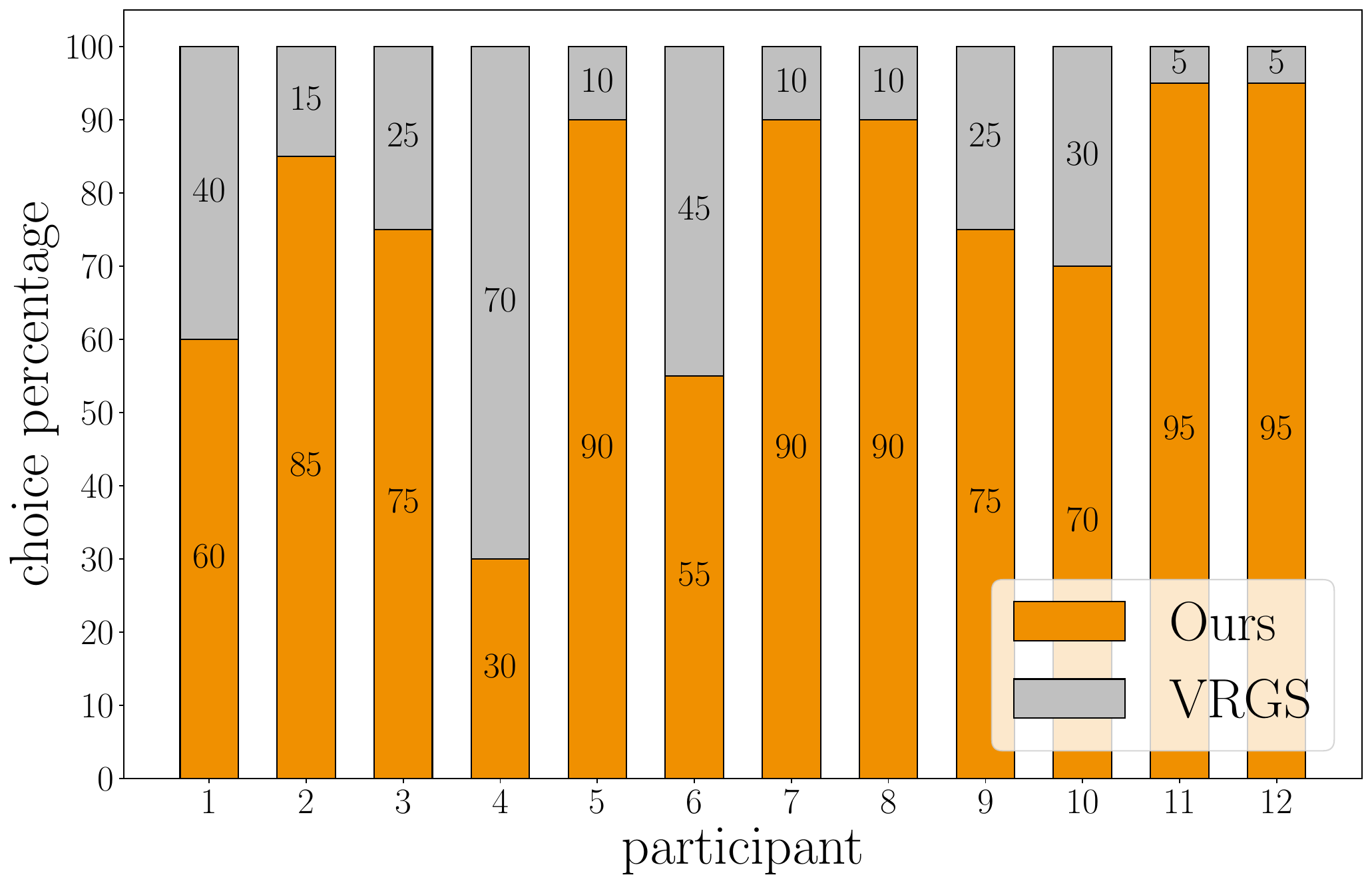}
    \caption{Results of our user study per participant, with the majority prefering our method.}
    \label{fig:user_study_results}
\end{figure}

\paragraph{Results.}
Participants in total made 240 decisions, their percentage distributions are shown in Fig.~\ref{fig:user_study_results}.
We observe a significantly higher ratio of participants who prefer ours to VRGS ($\sim76\%$ preferred ours, $SD=0.196$).
Binomial tests showed significance of this with $p<0.005$.

In the \textsc{garden} scene, our method was preferred overwhelmingly with over 90\%, as our method is able to showcase crisp details accurately (as seen in Fig.~\ref{fig:teaser}). The \textsc{playground} scene revealed the lowest preference for our method with 67\%, as here the initial viewpoint was further away, making details easier to miss for the participants.

\paragraph{Discussion.}
The high preference ratio for our method confirms the effectiveness of our method. 
This is underlined by the comments made by the participants, which highlighted the higher perceived quality and sharpness of our renderings.

\subsection{Ablation Studies}

\subsubsection{Performance Ablation}
We ablate our rendering performance in Tab.~\ref{tab:perf_ablation}, with individual features turned off.
Rendering time is evenly split between foveal and peripheral reconstruction (upper part), with peripheral rendering taking 4.9 ms on average.

The enhancements mentioned in Section~\ref{sec:method} all improve performance in a measurable way (middle part).
Opacity penalty for example increases render times by 2.5 ms, while depth culling slightly lowers peripheral render times at the cost of not culling neural points.

Our introduced quality enhancing features (lower parts) have performance impacts, especially the hierarchical sorting approach to fix popping~\cite{radl2024stopthepop}. However, as described in Sec.~\ref{sec:prototype}, it proved important to perceptual quality.

\begin{table}[]
\centering
\caption{\label{tab:perf_ablation}%
Performance Ablations of our features and improvements. \textit{Periphery GS} refers to the rasterization of the Gaussian splats and \textit{Fovea Points} to the rasterization of neural points. }
\small
\begin{tabular}{l|ccccc|r}
    &  Periphery GS & Fovea Points & CNN & Combine & Tonemapping & Combined \\ \hline
Ours                &   4.9 ms   &  3.5 ms      &  1.6 ms    &     0.5 ms   &   0.4 ms     & 10.9 ms\\\hline
w/o depth &    4.8 ms & 4.0 ms & 1.6 ms& 0.5 ms & 0.4 ms & +0.4 ms \\
w/o smaller neural net &   4.9 ms    &  3.5 ms     &  1.9 ms   &     0.5 ms   &   0.4 ms & +0.3 ms \\
w/o opacity penalty &  7.2 ms   &  3.7 ms     &  1.6 ms   &     0.5 ms   &   0.4 ms & +2.5 ms\\\hline
w/o popping fix &  2.6 ms   &  3.5 ms     &  1.6 ms   &     0.5 ms   &   0.4 ms & -2.3 ms\\
w/o combine $e$ &    4.9 ms    &  3.5 ms     &  1.6 ms   &     0.3 ms   &   0.4 ms     & -0.2 ms\\ 
\end{tabular}
\end{table}

\subsubsection{Densification Strategy for Gaussian Splatting}
Recently, Kheradmand et al.~\shortcite{kheradmand20243d} introduced a Markov chain-dependent Gaussian Splatting densification scheme, allowing for an exact determination of the number of Gaussians. 
This approach would be preferable since we need to maintain a conservative densification heuristic to prevent exceeding the VR limit.

\begin{figure}
    \centering
    \includegraphics[width=.99\linewidth]{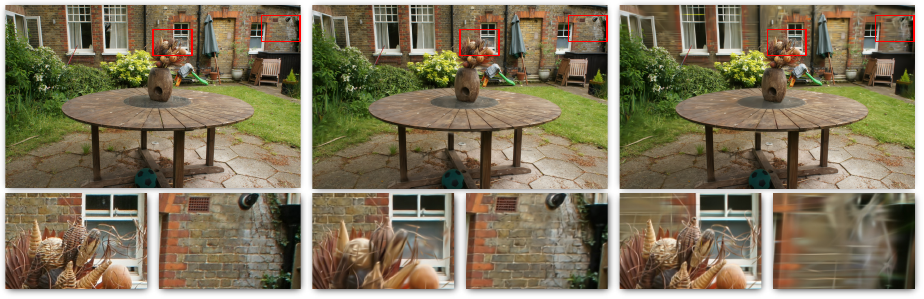}
    \begin{tabular}{*{3}{p{0.28\linewidth}<{\centering}}}
    \small (a) Ground Truth & \small(b) Ours (periphery) (0.4M) & \small(c) MCMC (0.5M)    
    \end{tabular}
    \caption{Densification strategies.
    3DGS-MCMC~\cite{kheradmand20243d} employ an densification scheme up to an exact number of primitives, however their background reconstruction quality is limited compared to ours.}
    \label{fig:densify_stategy}
\end{figure}

As shown in Fig.~\ref{fig:densify_stategy}, the low amount of primitives still result in great foreground reconstruction, however, deteriorated backgrounds.
This introduces severe artifacts in the peripheral reconstruction, including spiky Gaussians with high color variance, which is unsuitable for smooth peripheral rendering.
As such we use the heuristic densification strategy.

\section{Limitations \& Future Work}
We  inherit limitations from both Gaussian Splatting~\cite{kerbl3Dgaussians} and TRIPS~\cite{franke2024trips}. 
In particular, we require good capturing densities, as otherwise holes or missing areas are present. 
Furthermore, we are dependent on COLMAP for SfM/MVS results; otherwise, artifacts due to miscalibrations occur.
Slight camera pose errors are optimized by our pipeline; however, severe errors cannot be compensated for.

Our approach and the VRGS Gaussian baseline suffer from floaters which is further enhanced by the low densification thresholds used.
Passing through floaters can be perceived as flickering and is distracting in VR.
Stronger mitigation methods~\cite{philip2023floaters,huang20242d} would be beneficial in these cases.

Our hardware also presents challenges.
We had to increase the fovea sizes as the eye tracker was inaccurate during the experiments. 
Better calibrated and more accurate gaze position allow further performance increases with our method.

\section{Conclusion}

In conclusion, our proposed foveated rendering approach demonstrates a viable solution to overcome the computational demand for radiance field rendering in real-time virtual reality systems. 
By blending high-detail neural point rendering for the foveal region with smooth 3DGS rendering in the periphery, we obtain crisp rendering of radiance fields without compromising system performance.
Our hybrid solution meets the performance demands of interactive VR, offering a significant improvement in rendering quality and user experience, which might pave the way for more realistic and immersive VR applications such as video games or virtual tourism.

\begin{acks}
We appreciate the insightful feedback from the anonymous reviewers. Our gratitude also goes to Matthias Innmann, Stefan Romberg, Michael Gerstmayr, and Tim Habigt for the productive discussions. Thanks to Florian Hahlbohm for his excellent proofreading feedback.

Linus Franke was supported by the Bayerische Forschungsstiftung (Bavarian Research Foundation) AZ-1422-20 and the 5G innovation program of the German Federal Ministry for Digital and Transport under the funding code 165GU103B.
The authors gratefully acknowledge the scientific support and HPC resources provided by the Erlangen National High Performance Computing Center (NHR@FAU) of the Friedrich-Alexander-Universit\"at Erlangen-N\"urnberg (FAU) under the NHR project \textit{b162dc}. NHR funding is provided by federal and Bavarian state authorities. NHR@FAU hardware is partially funded by the German Research Foundation (DFG) – 440719683.
\end{acks}

\bibliographystyle{ACM-Reference-Format}
\bibliography{bib}


\begin{thebibliography}{129}


\ifx \showCODEN    \undefined \def \showCODEN     #1{\unskip}     \fi
\ifx \showDOI      \undefined \def \showDOI       #1{#1}\fi
\ifx \showISBNx    \undefined \def \showISBNx     #1{\unskip}     \fi
\ifx \showISBNxiii \undefined \def \showISBNxiii  #1{\unskip}     \fi
\ifx \showISSN     \undefined \def \showISSN      #1{\unskip}     \fi
\ifx \showLCCN     \undefined \def \showLCCN      #1{\unskip}     \fi
\ifx \shownote     \undefined \def \shownote      #1{#1}          \fi
\ifx \showarticletitle \undefined \def \showarticletitle #1{#1}   \fi
\ifx \showURL      \undefined \def \showURL       {\relax}        \fi
\providecommand\bibfield[2]{#2}
\providecommand\bibinfo[2]{#2}
\providecommand\natexlab[1]{#1}
\providecommand\showeprint[2][]{arXiv:#2}

\bibitem[Abou-Chakra et~al\mbox{.}(2024)]%
        {abou2024particlenerf}
\bibfield{author}{\bibinfo{person}{Jad Abou-Chakra}, \bibinfo{person}{Feras
  Dayoub}, {and} \bibinfo{person}{Niko S{\"u}nderhauf}.}
  \bibinfo{year}{2024}\natexlab{}.
\newblock \showarticletitle{ParticleNeRF: A Particle-Based Encoding for Online
  Neural Radiance Fields}. In \bibinfo{booktitle}{\emph{Proceedings of the
  IEEE/CVF Winter Conference on Applications of Computer Vision}}.
  \bibinfo{pages}{5975--5984}.
\newblock


\bibitem[Albert et~al\mbox{.}(2017)]%
        {albert2017latency}
\bibfield{author}{\bibinfo{person}{Rachel Albert}, \bibinfo{person}{Anjul
  Patney}, \bibinfo{person}{David Luebke}, {and} \bibinfo{person}{Joohwan
  Kim}.} \bibinfo{year}{2017}\natexlab{}.
\newblock \showarticletitle{{Latency Requirements for Foveated Rendering in
  Virtual Reality}}.
\newblock \bibinfo{journal}{\emph{ACM Transactions on Applied Perception
  (TAP)}} \bibinfo{volume}{14}, \bibinfo{number}{4} (\bibinfo{year}{2017}),
  \bibinfo{pages}{25}.
\newblock


\bibitem[Alexa et~al\mbox{.}(2004)]%
        {alexa2004point}
\bibfield{author}{\bibinfo{person}{Marc Alexa}, \bibinfo{person}{Markus Gross},
  \bibinfo{person}{Mark Pauly}, \bibinfo{person}{Hanspeter Pfister},
  \bibinfo{person}{Marc Stamminger}, {and} \bibinfo{person}{Matthias Zwicker}.}
  \bibinfo{year}{2004}\natexlab{}.
\newblock \showarticletitle{Point-based Computer Graphics}. In
  \bibinfo{booktitle}{\emph{ACM SIGGRAPH 2004 Course Notes}}.
  \bibinfo{pages}{7--es}.
\newblock


\bibitem[Aliev et~al\mbox{.}(2020)]%
        {aliev2020npbg}
\bibfield{author}{\bibinfo{person}{Kara-Ali Aliev}, \bibinfo{person}{Artem
  Sevastopolsky}, \bibinfo{person}{Maria Kolos}, \bibinfo{person}{Dmitry
  Ulyanov}, {and} \bibinfo{person}{Victor Lempitsky}.}
  \bibinfo{year}{2020}\natexlab{}.
\newblock \showarticletitle{Neural point-based graphics}. In
  \bibinfo{booktitle}{\emph{ECCV}}.
\newblock
\urldef\tempurl%
\url{https://doi.org/10.1007/978-3-030-58542-6_42}
\showDOI{\tempurl}


\bibitem[Barron et~al\mbox{.}(2021)]%
        {barron2021mipnerf}
\bibfield{author}{\bibinfo{person}{Jonathan~T. Barron}, \bibinfo{person}{Ben
  Mildenhall}, \bibinfo{person}{Matthew Tancik}, \bibinfo{person}{Peter
  Hedman}, \bibinfo{person}{Ricardo Martin-Brualla}, {and}
  \bibinfo{person}{Pratul~P. Srinivasan}.} \bibinfo{year}{2021}\natexlab{}.
\newblock \showarticletitle{Mip-NeRF: A Multiscale Representation for
  Anti-Aliasing Neural Radiance Fields}.
\newblock \bibinfo{journal}{\emph{ICCV}} (\bibinfo{year}{2021}).
\newblock
\urldef\tempurl%
\url{https://doi.org/10.1109/ICCV48922.2021.00580}
\showDOI{\tempurl}


\bibitem[Barron et~al\mbox{.}(2022)]%
        {barron2022mipnerf360}
\bibfield{author}{\bibinfo{person}{Jonathan~T. Barron}, \bibinfo{person}{Ben
  Mildenhall}, \bibinfo{person}{Dor Verbin}, \bibinfo{person}{Pratul~P.
  Srinivasan}, {and} \bibinfo{person}{Peter Hedman}.}
  \bibinfo{year}{2022}\natexlab{}.
\newblock \showarticletitle{Mip-NeRF 360: Unbounded Anti-Aliased Neural
  Radiance Fields}.
\newblock \bibinfo{journal}{\emph{CVPR}} (\bibinfo{year}{2022}).
\newblock
\urldef\tempurl%
\url{https://doi.org/10.1109/CVPR52688.2022.00539}
\showDOI{\tempurl}


\bibitem[Barron et~al\mbox{.}(2023)]%
        {barron2023ICCV}
\bibfield{author}{\bibinfo{person}{Jonathan~T. Barron}, \bibinfo{person}{Ben
  Mildenhall}, \bibinfo{person}{Dor Verbin}, \bibinfo{person}{Pratul~P.
  Srinivasan}, {and} \bibinfo{person}{Peter Hedman}.}
  \bibinfo{year}{2023}\natexlab{}.
\newblock \showarticletitle{Zip-NeRF: Anti-Aliased Grid-Based Neural Radiance
  Fields}. In \bibinfo{booktitle}{\emph{ICCV}}. \bibinfo{pages}{19640--19648}.
\newblock
\urldef\tempurl%
\url{https://doi.org/10.1109/ICCV51070.2023.01804}
\showDOI{\tempurl}


\bibitem[Bonopera et~al\mbox{.}(2020)]%
        {sibr2020}
\bibfield{author}{\bibinfo{person}{Sebastien Bonopera}, \bibinfo{person}{Peter
  Hedman}, \bibinfo{person}{Jerome Esnault}, \bibinfo{person}{Siddhant
  Prakash}, \bibinfo{person}{Simon Rodriguez}, \bibinfo{person}{Theo Thonat},
  \bibinfo{person}{Mehdi Benadel}, \bibinfo{person}{Gaurav Chaurasia},
  \bibinfo{person}{Julien Philip}, {and} \bibinfo{person}{George Drettakis}.}
  \bibinfo{year}{2020}\natexlab{}.
\newblock \bibinfo{title}{sibr: A System for Image Based Rendering}.
\newblock
\newblock
\urldef\tempurl%
\url{https://sibr.gitlabpages.inria.fr/}
\showURL{%
\tempurl}


\bibitem[Chaurasia et~al\mbox{.}(2013)]%
        {chaurasia2013depth}
\bibfield{author}{\bibinfo{person}{Gaurav Chaurasia}, \bibinfo{person}{Sylvain
  Duchene}, \bibinfo{person}{Olga Sorkine-Hornung}, {and}
  \bibinfo{person}{George Drettakis}.} \bibinfo{year}{2013}\natexlab{}.
\newblock \showarticletitle{Depth synthesis and local warps for plausible
  image-based navigation}.
\newblock \bibinfo{journal}{\emph{ACM TOG}} \bibinfo{volume}{32},
  \bibinfo{number}{3} (\bibinfo{year}{2013}), \bibinfo{pages}{1--12}.
\newblock


\bibitem[Chen et~al\mbox{.}(2022)]%
        {chen2022tensorf}
\bibfield{author}{\bibinfo{person}{Anpei Chen}, \bibinfo{person}{Zexiang Xu},
  \bibinfo{person}{Andreas Geiger}, \bibinfo{person}{Jingyi Yu}, {and}
  \bibinfo{person}{Hao Su}.} \bibinfo{year}{2022}\natexlab{}.
\newblock \showarticletitle{Tenso{RF}: Tensorial Radiance Fields}. In
  \bibinfo{booktitle}{\emph{ECCV}}.
\newblock
\urldef\tempurl%
\url{https://doi.org/10.1007/978-3-031-19824-3_20}
\showDOI{\tempurl}


\bibitem[Chen et~al\mbox{.}(2021)]%
        {chen2021MVSnerf}
\bibfield{author}{\bibinfo{person}{Anpei Chen}, \bibinfo{person}{Zexiang Xu},
  \bibinfo{person}{Fuqiang Zhao}, \bibinfo{person}{Xiaoshuai Zhang},
  \bibinfo{person}{Fanbo Xiang}, \bibinfo{person}{Jingyi Yu}, {and}
  \bibinfo{person}{Hao Su}.} \bibinfo{year}{2021}\natexlab{}.
\newblock \showarticletitle{Mvsnerf: Fast generalizable radiance field
  reconstruction from multi-view stereo}. In
  \bibinfo{booktitle}{\emph{Proceedings of the IEEE/CVF International
  Conference on Computer Vision}}. \bibinfo{pages}{14124--14133}.
\newblock


\bibitem[Chen and Wang(2024)]%
        {chen2024survey}
\bibfield{author}{\bibinfo{person}{Guikun Chen} {and} \bibinfo{person}{Wenguan
  Wang}.} \bibinfo{year}{2024}\natexlab{}.
\newblock \showarticletitle{A survey on 3d gaussian splatting}.
\newblock \bibinfo{journal}{\emph{arXiv preprint arXiv:2401.03890}}
  (\bibinfo{year}{2024}).
\newblock


\bibitem[Chen et~al\mbox{.}(2023)]%
        {chen2023mobilenerf}
\bibfield{author}{\bibinfo{person}{Zhiqin Chen}, \bibinfo{person}{Thomas
  Funkhouser}, \bibinfo{person}{Peter Hedman}, {and} \bibinfo{person}{Andrea
  Tagliasacchi}.} \bibinfo{year}{2023}\natexlab{}.
\newblock \showarticletitle{Mobilenerf: Exploiting the polygon rasterization
  pipeline for efficient neural field rendering on mobile architectures}. In
  \bibinfo{booktitle}{\emph{CVPR}}. \bibinfo{pages}{16569--16578}.
\newblock


\bibitem[Chibane et~al\mbox{.}(2021)]%
        {chibane2021stereo}
\bibfield{author}{\bibinfo{person}{Julian Chibane}, \bibinfo{person}{Aayush
  Bansal}, \bibinfo{person}{Verica Lazova}, {and} \bibinfo{person}{Gerard
  Pons-Moll}.} \bibinfo{year}{2021}\natexlab{}.
\newblock \showarticletitle{Stereo radiance fields (srf): Learning view
  synthesis for sparse views of novel scenes}. In
  \bibinfo{booktitle}{\emph{CVPR}}. \bibinfo{pages}{7911--7920}.
\newblock


\bibitem[Curcio et~al\mbox{.}(1990)]%
        {curcio1990human}
\bibfield{author}{\bibinfo{person}{Christine~A Curcio},
  \bibinfo{person}{Kenneth~R Sloan}, \bibinfo{person}{Robert~E Kalina}, {and}
  \bibinfo{person}{Anita~E Hendrickson}.} \bibinfo{year}{1990}\natexlab{}.
\newblock \showarticletitle{{Human Photoreceptor Topography}}.
\newblock \bibinfo{journal}{\emph{Journal of Comparative Neurology}}
  \bibinfo{volume}{292}, \bibinfo{number}{4} (\bibinfo{year}{1990}),
  \bibinfo{pages}{497--523}.
\newblock


\bibitem[Dai et~al\mbox{.}(2017)]%
        {dai2017bundlefusion}
\bibfield{author}{\bibinfo{person}{Angela Dai}, \bibinfo{person}{Matthias
  Nie{\ss}ner}, \bibinfo{person}{Michael Zollh{\"o}fer},
  \bibinfo{person}{Shahram Izadi}, {and} \bibinfo{person}{Christian Theobalt}.}
  \bibinfo{year}{2017}\natexlab{}.
\newblock \showarticletitle{Bundlefusion: Real-time globally consistent 3d
  reconstruction using on-the-fly surface reintegration}.
\newblock \bibinfo{journal}{\emph{ACM Transactions on Graphics (ToG)}}
  \bibinfo{volume}{36}, \bibinfo{number}{4} (\bibinfo{year}{2017}),
  \bibinfo{pages}{1}.
\newblock


\bibitem[Debevec et~al\mbox{.}(1998)]%
        {debevec1998efficient}
\bibfield{author}{\bibinfo{person}{Paul Debevec}, \bibinfo{person}{Yizhou Yu},
  {and} \bibinfo{person}{George Boshokov}.} \bibinfo{year}{1998}\natexlab{}.
\newblock \showarticletitle{Efficient view-dependent IBR with projective
  texture-mapping}. In \bibinfo{booktitle}{\emph{EG Rendering Workshop}},
  Vol.~\bibinfo{volume}{4}.
\newblock


\bibitem[Deng et~al\mbox{.}(2022)]%
        {deng2022fovnerf}
\bibfield{author}{\bibinfo{person}{Nianchen Deng}, \bibinfo{person}{Zhenyi He},
  \bibinfo{person}{Jiannan Ye}, \bibinfo{person}{Budmonde Duinkharjav},
  \bibinfo{person}{Praneeth Chakravarthula}, \bibinfo{person}{Xubo Yang}, {and}
  \bibinfo{person}{Qi Sun}.} \bibinfo{year}{2022}\natexlab{}.
\newblock \showarticletitle{FoV-NeRF: Foveated Neural Radiance Fields for
  Virtual Reality}.
\newblock \bibinfo{journal}{\emph{IEEE TVCG}} (\bibinfo{year}{2022}).
\newblock
\urldef\tempurl%
\url{https://doi.org/10.1109/TVCG.2022.3203102}
\showDOI{\tempurl}


\bibitem[Duckworth et~al\mbox{.}(2024)]%
        {duckworth2023smerf}
\bibfield{author}{\bibinfo{person}{Daniel Duckworth}, \bibinfo{person}{Peter
  Hedman}, \bibinfo{person}{Christian Reiser}, \bibinfo{person}{Peter Zhizhin},
  \bibinfo{person}{Jean-Fran{\c{c}}ois Thibert}, \bibinfo{person}{Mario
  Lu{\v{c}}i{\'c}}, \bibinfo{person}{Richard Szeliski}, {and}
  \bibinfo{person}{Jonathan~T Barron}.} \bibinfo{year}{2024}\natexlab{}.
\newblock \showarticletitle{Smerf: Streamable memory efficient radiance fields
  for real-time large-scene exploration}.
\newblock \bibinfo{journal}{\emph{ACM Transactions on Graphics (TOG)}}
  \bibinfo{volume}{43}, \bibinfo{number}{4} (\bibinfo{year}{2024}),
  \bibinfo{pages}{1--13}.
\newblock


\bibitem[Eisemann et~al\mbox{.}(2008)]%
        {eisemann2008floating}
\bibfield{author}{\bibinfo{person}{Martin Eisemann}, \bibinfo{person}{Bert
  De~Decker}, \bibinfo{person}{Marcus Magnor}, \bibinfo{person}{Philippe
  Bekaert}, \bibinfo{person}{Edilson De~Aguiar}, \bibinfo{person}{Naveed
  Ahmed}, \bibinfo{person}{Christian Theobalt}, {and} \bibinfo{person}{Anita
  Sellent}.} \bibinfo{year}{2008}\natexlab{}.
\newblock \showarticletitle{Floating textures}. In
  \bibinfo{booktitle}{\emph{Comput. Graph. Forum}}, Vol.~\bibinfo{volume}{27}.
  Wiley Online Library, \bibinfo{pages}{409--418}.
\newblock


\bibitem[Franke et~al\mbox{.}(2021)]%
        {franke2021time}
\bibfield{author}{\bibinfo{person}{Linus Franke}, \bibinfo{person}{Laura Fink},
  \bibinfo{person}{Jana Martschinke}, \bibinfo{person}{Kai Selgrad}, {and}
  \bibinfo{person}{Marc Stamminger}.} \bibinfo{year}{2021}\natexlab{}.
\newblock \showarticletitle{Time-Warped Foveated Rendering for Virtual Reality
  Headsets}.
\newblock \bibinfo{journal}{\emph{Computer Graphics Forum}}
  \bibinfo{volume}{40}, \bibinfo{number}{1} (\bibinfo{year}{2021}),
  \bibinfo{pages}{110--123}.
\newblock
\urldef\tempurl%
\url{https://doi.org/10.1111/cgf.14176}
\showDOI{\tempurl}
\showeprint{https://onlinelibrary.wiley.com/doi/pdf/10.1111/cgf.14176}


\bibitem[Franke et~al\mbox{.}(2023)]%
        {franke2023vet}
\bibfield{author}{\bibinfo{person}{Linus Franke}, \bibinfo{person}{Darius
  R{\"u}ckert}, \bibinfo{person}{Laura Fink}, \bibinfo{person}{Matthias
  Innmann}, {and} \bibinfo{person}{Marc Stamminger}.}
  \bibinfo{year}{2023}\natexlab{}.
\newblock \showarticletitle{VET: Visual Error Tomography for Point Cloud
  Completion and High-Quality Neural Rendering}. In
  \bibinfo{booktitle}{\emph{SIGGRAPH Asia}}. \bibinfo{publisher}{Association
  for Computing Machinery}, \bibinfo{address}{New York, NY, USA}.
\newblock


\bibitem[Franke et~al\mbox{.}(2024)]%
        {franke2024trips}
\bibfield{author}{\bibinfo{person}{Linus Franke}, \bibinfo{person}{Darius
  R{\"u}ckert}, \bibinfo{person}{Laura Fink}, {and} \bibinfo{person}{Marc
  Stamminger}.} \bibinfo{year}{2024}\natexlab{}.
\newblock \showarticletitle{TRIPS: Trilinear Point Splatting for Real-Time
  Radiance Field Rendering}.
\newblock \bibinfo{journal}{\emph{Comput. Graph. Forum}} \bibinfo{volume}{43},
  \bibinfo{number}{2} (\bibinfo{year}{2024}).
\newblock
\urldef\tempurl%
\url{https://doi.org/10.1111/cgf.15012}
\showDOI{\tempurl}


\bibitem[Fridovich-Keil et~al\mbox{.}(2022)]%
        {Fridovich2022}
\bibfield{author}{\bibinfo{person}{Sara Fridovich-Keil}, \bibinfo{person}{Alex
  Yu}, \bibinfo{person}{Matthew Tancik}, \bibinfo{person}{Qinhong Chen},
  \bibinfo{person}{Benjamin Recht}, {and} \bibinfo{person}{Angjoo Kanazawa}.}
  \bibinfo{year}{2022}\natexlab{}.
\newblock \showarticletitle{Plenoxels: Radiance Fields without Neural
  Networks}. In \bibinfo{booktitle}{\emph{2022 IEEE/CVF Conference on Computer
  Vision and Pattern Recognition (CVPR)}}. \bibinfo{pages}{5491--5500}.
\newblock
\urldef\tempurl%
\url{https://doi.org/10.1109/CVPR52688.2022.00542}
\showDOI{\tempurl}


\bibitem[Friston et~al\mbox{.}(2019)]%
        {friston2019perceptual}
\bibfield{author}{\bibinfo{person}{Sebastian Friston}, \bibinfo{person}{Tobias
  Ritschel}, {and} \bibinfo{person}{Anthony Steed}.}
  \bibinfo{year}{2019}\natexlab{}.
\newblock \showarticletitle{Perceptual rasterization for head-mounted display
  image synthesis}.
\newblock \bibinfo{journal}{\emph{ACM Transactions on Graphics (TOG)}}
  \bibinfo{volume}{38}, \bibinfo{number}{4} (\bibinfo{year}{2019}),
  \bibinfo{pages}{1--14}.
\newblock


\bibitem[Goesele et~al\mbox{.}(2007)]%
        {goesele2007multi}
\bibfield{author}{\bibinfo{person}{Michael Goesele}, \bibinfo{person}{Noah
  Snavely}, \bibinfo{person}{Brian Curless}, \bibinfo{person}{Hugues Hoppe},
  {and} \bibinfo{person}{Steven~M Seitz}.} \bibinfo{year}{2007}\natexlab{}.
\newblock \showarticletitle{Multi-view stereo for community photo collections}.
  In \bibinfo{booktitle}{\emph{2007 IEEE 11th International Conference on
  Computer Vision}}. IEEE, \bibinfo{pages}{1--8}.
\newblock


\bibitem[Goldstein and Brockmole(2016)]%
        {goldstein2016sensation}
\bibfield{author}{\bibinfo{person}{E~Bruce Goldstein} {and}
  \bibinfo{person}{James Brockmole}.} \bibinfo{year}{2016}\natexlab{}.
\newblock \bibinfo{booktitle}{\emph{{Sensation and Perception}}}.
\newblock \bibinfo{publisher}{Cengage Learning}.
\newblock


\bibitem[Gortler et~al\mbox{.}(1996)]%
        {gortler1996lumigraph}
\bibfield{author}{\bibinfo{person}{Steven~J. Gortler}, \bibinfo{person}{Radek
  Grzeszczuk}, \bibinfo{person}{Richard Szeliski}, {and}
  \bibinfo{person}{Michael~F. Cohen}.} \bibinfo{year}{1996}\natexlab{}.
\newblock \showarticletitle{The {L}umigraph}. In
  \bibinfo{booktitle}{\emph{CGIT}}.
\newblock
\urldef\tempurl%
\url{https://doi.org/10.1145/237170.237200}
\showDOI{\tempurl}


\bibitem[Guenter et~al\mbox{.}(2012)]%
        {guenter2012foveated}
\bibfield{author}{\bibinfo{person}{Brian Guenter}, \bibinfo{person}{Mark
  Finch}, \bibinfo{person}{Steven Drucker}, \bibinfo{person}{Desney Tan}, {and}
  \bibinfo{person}{John Snyder}.} \bibinfo{year}{2012}\natexlab{}.
\newblock \showarticletitle{{Foveated 3D Graphics}}.
\newblock \bibinfo{journal}{\emph{ACM Transactions on Graphics (TOG)}}
  \bibinfo{volume}{31}, \bibinfo{number}{6} (\bibinfo{year}{2012}),
  \bibinfo{pages}{164}.
\newblock


\bibitem[Hahlbohm et~al\mbox{.}(2025a)]%
        {hahlbohm2024inpc}
\bibfield{author}{\bibinfo{person}{Florian Hahlbohm}, \bibinfo{person}{Linus
  Franke}, \bibinfo{person}{Moritz Kappel}, \bibinfo{person}{Susana Castillo},
  \bibinfo{person}{Martin Eisemann}, \bibinfo{person}{Marc Stamminger}, {and}
  \bibinfo{person}{Marcus Magnor}.} \bibinfo{year}{2025}\natexlab{a}.
\newblock \showarticletitle{INPC: Implicit Neural Point Clouds for Radiance
  Field Rendering}. In \bibinfo{booktitle}{\emph{2025 International Conference
  on 3D Vision (3DV)}}.
\newblock
\urldef\tempurl%
\url{https://fhahlbohm.github.io/inpc/}
\showURL{%
\tempurl}


\bibitem[Hahlbohm et~al\mbox{.}(2025b)]%
        {hahlbohm2025htgs}
\bibfield{author}{\bibinfo{person}{Florian Hahlbohm}, \bibinfo{person}{Fabian
  Friederichs}, \bibinfo{person}{Tim Weyrich}, \bibinfo{person}{Linus Franke},
  \bibinfo{person}{Moritz Kappel}, \bibinfo{person}{Susana Castillo},
  \bibinfo{person}{Marc Stamminger}, \bibinfo{person}{Martin Eisemann}, {and}
  \bibinfo{person}{Marcus Magnor}.} \bibinfo{year}{2025}\natexlab{b}.
\newblock \showarticletitle{Efficient Perspective-Correct 3D Gaussian Splatting
  Using Hybrid Transparency}.
\newblock \bibinfo{journal}{\emph{Computer Graphics Forum}}
  \bibinfo{volume}{44}, \bibinfo{number}{2} (\bibinfo{year}{2025}).
\newblock
\urldef\tempurl%
\url{https://fhahlbohm.github.io/htgs/}
\showURL{%
\tempurl}


\bibitem[Hahlbohm et~al\mbox{.}(2023)]%
        {hahlbohm2023plenopticpoints}
\bibfield{author}{\bibinfo{person}{Florian Hahlbohm}, \bibinfo{person}{Moritz
  Kappel}, \bibinfo{person}{Jan-Philipp Tauscher}, \bibinfo{person}{Martin
  Eisemann}, {and} \bibinfo{person}{Marcus Magnor}.}
  \bibinfo{year}{2023}\natexlab{}.
\newblock \showarticletitle{PlenopticPoints: Rasterizing Neural Feature Points
  for High-Quality Novel View Synthesis}. In \bibinfo{booktitle}{\emph{Proc.
  Vision, Modeling and Visualization ({VMV})}},
  \bibfield{editor}{\bibinfo{person}{T.~Grosch} {and}
  \bibinfo{person}{M.~Guthe}} (Eds.). Eurographics, \bibinfo{pages}{53--61}.
\newblock
\showISBNx{978-3-03868-232-5}
\urldef\tempurl%
\url{https://doi.org/10.2312/vmv.20231226}
\showDOI{\tempurl}


\bibitem[Harrer et~al\mbox{.}(2023)]%
        {harrerfranke2023inovis}
\bibfield{author}{\bibinfo{person}{Mathias Harrer}, \bibinfo{person}{Linus
  Franke}, \bibinfo{person}{Laura Fink}, \bibinfo{person}{Marc Stamminger},
  {and} \bibinfo{person}{Tim Weyrich}.} \bibinfo{year}{2023}\natexlab{}.
\newblock \showarticletitle{Inovis: Instant Novel-View Synthesis}. In
  \bibinfo{booktitle}{\emph{SIGGRAPH Asia}} (Sydney, NSW, Australia, December
  12-15, 2023). \bibinfo{publisher}{Association for Computing Machinery},
  \bibinfo{address}{New York, NY, USA}, \bibinfo{numpages}{12}~pages.
\newblock
\urldef\tempurl%
\url{https://doi.org/10.1145/3610548.3618216}
\showDOI{\tempurl}


\bibitem[Hedman et~al\mbox{.}(2018)]%
        {hedman2018deep}
\bibfield{author}{\bibinfo{person}{Peter Hedman}, \bibinfo{person}{Julien
  Philip}, \bibinfo{person}{True Price}, \bibinfo{person}{Jan-Michael Frahm},
  \bibinfo{person}{George Drettakis}, {and} \bibinfo{person}{Gabriel Brostow}.}
  \bibinfo{year}{2018}\natexlab{}.
\newblock \showarticletitle{Deep blending for free-viewpoint image-based
  rendering}.
\newblock \bibinfo{journal}{\emph{ACM TOG}} \bibinfo{volume}{37},
  \bibinfo{number}{6} (\bibinfo{year}{2018}), \bibinfo{pages}{1--15}.
\newblock


\bibitem[Hoffman et~al\mbox{.}(2018)]%
        {hoffman2018limits}
\bibfield{author}{\bibinfo{person}{David Hoffman}, \bibinfo{person}{Zoe Meraz},
  {and} \bibinfo{person}{Eric Turner}.} \bibinfo{year}{2018}\natexlab{}.
\newblock \showarticletitle{Limits of peripheral acuity and implications for VR
  system design}.
\newblock \bibinfo{journal}{\emph{Journal of the Society for Information
  Display}} \bibinfo{volume}{26}, \bibinfo{number}{8} (\bibinfo{year}{2018}),
  \bibinfo{pages}{483--495}.
\newblock


\bibitem[Huang et~al\mbox{.}(2024a)]%
        {Huang2DGS2024}
\bibfield{author}{\bibinfo{person}{Binbin Huang}, \bibinfo{person}{Zehao Yu},
  \bibinfo{person}{Anpei Chen}, \bibinfo{person}{Andreas Geiger}, {and}
  \bibinfo{person}{Shenghua Gao}.} \bibinfo{year}{2024}\natexlab{a}.
\newblock \showarticletitle{2D Gaussian Splatting for Geometrically Accurate
  Radiance Fields}. In \bibinfo{booktitle}{\emph{SIGGRAPH 2024 Conference
  Papers}}. \bibinfo{publisher}{Association for Computing Machinery}.
\newblock
\urldef\tempurl%
\url{https://doi.org/10.1145/3641519.3657428}
\showDOI{\tempurl}


\bibitem[Huang et~al\mbox{.}(2024b)]%
        {huang20242d}
\bibfield{author}{\bibinfo{person}{Binbin Huang}, \bibinfo{person}{Zehao Yu},
  \bibinfo{person}{Anpei Chen}, \bibinfo{person}{Andreas Geiger}, {and}
  \bibinfo{person}{Shenghua Gao}.} \bibinfo{year}{2024}\natexlab{b}.
\newblock \showarticletitle{2d gaussian splatting for geometrically accurate
  radiance fields}. In \bibinfo{booktitle}{\emph{SIGGRAPH}}.
  \bibinfo{pages}{1--11}.
\newblock


\bibitem[Huang et~al\mbox{.}(2023)]%
        {huang2023sc}
\bibfield{author}{\bibinfo{person}{Yi-Hua Huang}, \bibinfo{person}{Yang-Tian
  Sun}, \bibinfo{person}{Ziyi Yang}, \bibinfo{person}{Xiaoyang Lyu},
  \bibinfo{person}{Yan-Pei Cao}, {and} \bibinfo{person}{Xiaojuan Qi}.}
  \bibinfo{year}{2023}\natexlab{}.
\newblock \showarticletitle{SC-GS: Sparse-Controlled Gaussian Splatting for
  Editable Dynamic Scenes}.
\newblock \bibinfo{journal}{\emph{arXiv preprint arXiv:2312.14937}}
  (\bibinfo{year}{2023}).
\newblock


\bibitem[Jabbireddy et~al\mbox{.}(2022)]%
        {jabbireddy2022foveated}
\bibfield{author}{\bibinfo{person}{Susmija Jabbireddy},
  \bibinfo{person}{Xuetong Sun}, \bibinfo{person}{Xiaoxu Meng}, {and}
  \bibinfo{person}{Amitabh Varshney}.} \bibinfo{year}{2022}\natexlab{}.
\newblock \showarticletitle{Foveated rendering: Motivation, taxonomy, and
  research directions}.
\newblock \bibinfo{journal}{\emph{arXiv preprint arXiv:2205.04529}}
  (\bibinfo{year}{2022}).
\newblock


\bibitem[Jiang et~al\mbox{.}(2024)]%
        {jiang2024vr}
\bibfield{author}{\bibinfo{person}{Ying Jiang}, \bibinfo{person}{Chang Yu},
  \bibinfo{person}{Tianyi Xie}, \bibinfo{person}{Xuan Li},
  \bibinfo{person}{Yutao Feng}, \bibinfo{person}{Huamin Wang},
  \bibinfo{person}{Minchen Li}, \bibinfo{person}{Henry Lau},
  \bibinfo{person}{Feng Gao}, \bibinfo{person}{Yin Yang}, {et~al\mbox{.}}}
  \bibinfo{year}{2024}\natexlab{}.
\newblock \showarticletitle{VR-GS: a physical dynamics-aware interactive
  gaussian splatting system in virtual reality}. In
  \bibinfo{booktitle}{\emph{ACM SIGGRAPH 2024 Conference Papers}}.
  \bibinfo{pages}{1--1}.
\newblock


\bibitem[Kaplanyan et~al\mbox{.}(2019)]%
        {kaplanyan2019deepfovea}
\bibfield{author}{\bibinfo{person}{Anton~S Kaplanyan}, \bibinfo{person}{Anton
  Sochenov}, \bibinfo{person}{Thomas Leimk{\"u}hler}, \bibinfo{person}{Mikhail
  Okunev}, \bibinfo{person}{Todd Goodall}, {and} \bibinfo{person}{Gizem Rufo}.}
  \bibinfo{year}{2019}\natexlab{}.
\newblock \showarticletitle{DeepFovea: Neural reconstruction for foveated
  rendering and video compression using learned statistics of natural videos}.
\newblock \bibinfo{journal}{\emph{ACM Transactions on Graphics (TOG)}}
  \bibinfo{volume}{38}, \bibinfo{number}{6} (\bibinfo{year}{2019}),
  \bibinfo{pages}{1--13}.
\newblock


\bibitem[Kappel et~al\mbox{.}(2024)]%
        {kappel2024d}
\bibfield{author}{\bibinfo{person}{Moritz Kappel}, \bibinfo{person}{Florian
  Hahlbohm}, \bibinfo{person}{Timon Scholz}, \bibinfo{person}{Susana Castillo},
  \bibinfo{person}{Christian Theobalt}, \bibinfo{person}{Martin Eisemann},
  \bibinfo{person}{Vladislav Golyanik}, {and} \bibinfo{person}{Marcus Magnor}.}
  \bibinfo{year}{2024}\natexlab{}.
\newblock \showarticletitle{D-NPC: Dynamic Neural Point Clouds for Non-Rigid
  View Synthesis from Monocular Video}.
\newblock \bibinfo{journal}{\emph{arXiv preprint arXiv:2406.10078}}
  (\bibinfo{year}{2024}).
\newblock


\bibitem[Keller et~al\mbox{.}(2013)]%
        {keller2013realtime}
\bibfield{author}{\bibinfo{person}{Maik Keller}, \bibinfo{person}{Damien
  Lefloch}, \bibinfo{person}{Martin Lambers}, \bibinfo{person}{Shahram Izadi},
  \bibinfo{person}{Tim Weyrich}, {and} \bibinfo{person}{Andreas Kolb}.}
  \bibinfo{year}{2013}\natexlab{}.
\newblock \showarticletitle{Real-time {3D} Reconstruction in Dynamic Scenes
  using Point-based Fusion}. In \bibinfo{booktitle}{\emph{Proc. of Joint
  3DIM/3DPVT Conference (3DV)}}. \bibinfo{pages}{1--8}.
\newblock
\newblock
\shownote{Selected for oral presentation.}.


\bibitem[Kerbl et~al\mbox{.}(2023)]%
        {kerbl3Dgaussians}
\bibfield{author}{\bibinfo{person}{Bernhard Kerbl}, \bibinfo{person}{Georgios
  Kopanas}, \bibinfo{person}{Thomas Leimk{\"u}hler}, {and}
  \bibinfo{person}{George Drettakis}.} \bibinfo{year}{2023}\natexlab{}.
\newblock \showarticletitle{{3D Gaussian} Splatting for Real-Time Radiance
  Field Rendering}.
\newblock \bibinfo{journal}{\emph{ACM TOG}} \bibinfo{volume}{42},
  \bibinfo{number}{4} (\bibinfo{date}{July} \bibinfo{year}{2023}).
\newblock
\urldef\tempurl%
\url{https://doi.org/10.1145/3592433}
\showDOI{\tempurl}


\bibitem[Kerbl et~al\mbox{.}(2024)]%
        {kerbl2024hierarchical}
\bibfield{author}{\bibinfo{person}{Bernhard Kerbl}, \bibinfo{person}{Andreas
  Meuleman}, \bibinfo{person}{Georgios Kopanas}, \bibinfo{person}{Michael
  Wimmer}, \bibinfo{person}{Alexandre Lanvin}, {and} \bibinfo{person}{George
  Drettakis}.} \bibinfo{year}{2024}\natexlab{}.
\newblock \showarticletitle{A hierarchical 3d gaussian representation for
  real-time rendering of very large datasets}.
\newblock \bibinfo{journal}{\emph{ACM TOG}} \bibinfo{volume}{43},
  \bibinfo{number}{4} (\bibinfo{year}{2024}), \bibinfo{pages}{1--15}.
\newblock


\bibitem[Kheradmand et~al\mbox{.}(2024)]%
        {kheradmand20243d}
\bibfield{author}{\bibinfo{person}{Shakiba Kheradmand}, \bibinfo{person}{Daniel
  Rebain}, \bibinfo{person}{Gopal Sharma}, \bibinfo{person}{Weiwei Sun},
  \bibinfo{person}{Jeff Tseng}, \bibinfo{person}{Hossam Isack},
  \bibinfo{person}{Abhishek Kar}, \bibinfo{person}{Andrea Tagliasacchi}, {and}
  \bibinfo{person}{Kwang~Moo Yi}.} \bibinfo{year}{2024}\natexlab{}.
\newblock \showarticletitle{3D Gaussian Splatting as Markov Chain Monte Carlo}.
\newblock \bibinfo{journal}{\emph{arXiv preprint arXiv:2404.09591}}
  (\bibinfo{year}{2024}).
\newblock


\bibitem[Kim et~al\mbox{.}(2019)]%
        {kim2019foveated}
\bibfield{author}{\bibinfo{person}{Jonghyun Kim}, \bibinfo{person}{Youngmo
  Jeong}, \bibinfo{person}{Michael Stengel}, \bibinfo{person}{Kaan Aksit},
  \bibinfo{person}{Rachel~A Albert}, \bibinfo{person}{Ben Boudaoud},
  \bibinfo{person}{Trey Greer}, \bibinfo{person}{Joohwan Kim},
  \bibinfo{person}{Ward Lopes}, \bibinfo{person}{Zander Majercik},
  {et~al\mbox{.}}} \bibinfo{year}{2019}\natexlab{}.
\newblock \showarticletitle{Foveated AR: dynamically-foveated augmented reality
  display.}
\newblock \bibinfo{journal}{\emph{ACM Trans. Graph.}} \bibinfo{volume}{38},
  \bibinfo{number}{4} (\bibinfo{year}{2019}), \bibinfo{pages}{99--1}.
\newblock


\bibitem[Knapitsch et~al\mbox{.}(2017)]%
        {Knapitsch2017}
\bibfield{author}{\bibinfo{person}{Arno Knapitsch}, \bibinfo{person}{Jaesik
  Park}, \bibinfo{person}{Qian-Yi Zhou}, {and} \bibinfo{person}{Vladlen
  Koltun}.} \bibinfo{year}{2017}\natexlab{}.
\newblock \showarticletitle{Tanks and Temples: Benchmarking Large-Scale Scene
  Reconstruction}.
\newblock \bibinfo{journal}{\emph{ACM Transactions on Graphics}}
  \bibinfo{volume}{36}, \bibinfo{number}{4} (\bibinfo{year}{2017}).
\newblock


\bibitem[Kobbelt and Botsch(2004)]%
        {kobbelt2004survey}
\bibfield{author}{\bibinfo{person}{Leif Kobbelt} {and} \bibinfo{person}{Mario
  Botsch}.} \bibinfo{year}{2004}\natexlab{}.
\newblock \showarticletitle{A survey of point-based techniques in computer
  graphics}.
\newblock \bibinfo{journal}{\emph{Computers \& Graphics}} \bibinfo{volume}{28},
  \bibinfo{number}{6} (\bibinfo{year}{2004}), \bibinfo{pages}{801--814}.
\newblock


\bibitem[Kopanas and Drettakis(2023)]%
        {kopanas2023improving}
\bibfield{author}{\bibinfo{person}{Georgios Kopanas} {and}
  \bibinfo{person}{George Drettakis}.} \bibinfo{year}{2023}\natexlab{}.
\newblock \showarticletitle{{Improving NeRF Quality by Progressive Camera
  Placement for Free-Viewpoint Navigation}}. In
  \bibinfo{booktitle}{\emph{Vision, Modeling, and Visualization}},
  \bibfield{editor}{\bibinfo{person}{Michael Guthe} {and}
  \bibinfo{person}{Thorsten Grosch}} (Eds.). \bibinfo{publisher}{The
  Eurographics Association}.
\newblock
\showISBNx{978-3-03868-232-5}
\urldef\tempurl%
\url{https://doi.org/10.2312/vmv.20231222}
\showDOI{\tempurl}


\bibitem[Kopanas et~al\mbox{.}(2022)]%
        {kopanas2022catacaustics}
\bibfield{author}{\bibinfo{person}{Georgios Kopanas}, \bibinfo{person}{Thomas
  Leimk{\"u}hler}, \bibinfo{person}{Gilles Rainer},
  \bibinfo{person}{Cl{\'e}ment Jambon}, {and} \bibinfo{person}{George
  Drettakis}.} \bibinfo{year}{2022}\natexlab{}.
\newblock \showarticletitle{Neural Point Catacaustics for Novel-View Synthesis
  of Reflections}.
\newblock \bibinfo{journal}{\emph{ACM TOG}} (\bibinfo{year}{2022}).
\newblock
\urldef\tempurl%
\url{https://doi.org/10.1145/3550454.3555497}
\showDOI{\tempurl}


\bibitem[Kopanas et~al\mbox{.}(2021)]%
        {kopanas2021perviewopt}
\bibfield{author}{\bibinfo{person}{Georgios Kopanas}, \bibinfo{person}{Julien
  Philip}, \bibinfo{person}{Thomas Leimkühler}, {and} \bibinfo{person}{George
  Drettakis}.} \bibinfo{year}{2021}\natexlab{}.
\newblock \showarticletitle{Point-Based Neural Rendering with Per-View
  Optimization}.
\newblock \bibinfo{journal}{\emph{CGF}} (\bibinfo{year}{2021}).
\newblock
\urldef\tempurl%
\url{https://doi.org/10.1111/cgf.14339}
\showDOI{\tempurl}


\bibitem[Koskela et~al\mbox{.}(2016)]%
        {koskela2016foveated}
\bibfield{author}{\bibinfo{person}{Matias Koskela}, \bibinfo{person}{Timo
  Viitanen}, \bibinfo{person}{Pekka J{\"a}{\"a}skel{\"a}inen}, {and}
  \bibinfo{person}{Jarmo Takala}.} \bibinfo{year}{2016}\natexlab{}.
\newblock \showarticletitle{Foveated path tracing: a literature review and a
  performance gain analysis}. In \bibinfo{booktitle}{\emph{Advances in Visual
  Computing: 12th International Symposium, ISVC 2016, Las Vegas, NV, USA,
  December 12-14, 2016, Proceedings, Part I 12}}. Springer,
  \bibinfo{pages}{723--732}.
\newblock


\bibitem[Kreylos(2018)]%
        {kreylos2018displayresolution}
\bibfield{author}{\bibinfo{person}{Oliver Kreylos}.}
  \bibinfo{year}{2018}\natexlab{}.
\newblock \bibinfo{title}{The Display Resolution of Head-mounted Displays}.
\newblock
  \bibinfo{howpublished}{\url{https://web.archive.org/web/20240723174740/http://doc-ok.org/?p=1677}}.
\newblock


\bibitem[Kreylos(2019)]%
        {kreylos2019vivepro}
\bibfield{author}{\bibinfo{person}{Oliver Kreylos}.}
  \bibinfo{year}{2019}\natexlab{}.
\newblock \bibinfo{title}{Field of View and Resolution of the PlayStation VR
  Headset}.
\newblock
  \bibinfo{howpublished}{\url{https://web.archive.org/web/20240614024653/https://doc-ok.org/?p=1882}}.
\newblock


\bibitem[Ledig et~al\mbox{.}(2017)]%
        {vgg19loss}
\bibfield{author}{\bibinfo{person}{Christian Ledig}, \bibinfo{person}{Lucas
  Theis}, \bibinfo{person}{Ferenc Husz{\'a}r}, \bibinfo{person}{Jose
  Caballero}, \bibinfo{person}{Andrew Cunningham}, \bibinfo{person}{Alejandro
  Acosta}, \bibinfo{person}{Andrew Aitken}, \bibinfo{person}{Alykhan Tejani},
  \bibinfo{person}{Johannes Totz}, \bibinfo{person}{Zehan Wang},
  {et~al\mbox{.}}} \bibinfo{year}{2017}\natexlab{}.
\newblock \showarticletitle{Photo-realistic single image super-resolution using
  a generative adversarial network}. In \bibinfo{booktitle}{\emph{CVPR}}.
  \bibinfo{pages}{4681--4690}.
\newblock


\bibitem[Li et~al\mbox{.}(2022)]%
        {li2022immersive}
\bibfield{author}{\bibinfo{person}{Ke Li}, \bibinfo{person}{Tim Rolff},
  \bibinfo{person}{Susanne Schmidt}, \bibinfo{person}{Reinhard Bacher},
  \bibinfo{person}{Simone Frintrop}, \bibinfo{person}{Wim Leemans}, {and}
  \bibinfo{person}{Frank Steinicke}.} \bibinfo{year}{2022}\natexlab{}.
\newblock \showarticletitle{Immersive neural graphics primitives}.
\newblock \bibinfo{journal}{\emph{arXiv preprint arXiv:2211.13494}}
  (\bibinfo{year}{2022}).
\newblock


\bibitem[Liao et~al\mbox{.}(2022)]%
        {liao2022kitti}
\bibfield{author}{\bibinfo{person}{Yiyi Liao}, \bibinfo{person}{Jun Xie}, {and}
  \bibinfo{person}{Andreas Geiger}.} \bibinfo{year}{2022}\natexlab{}.
\newblock \showarticletitle{KITTI-360: A novel dataset and benchmarks for urban
  scene understanding in 2d and 3d}.
\newblock \bibinfo{journal}{\emph{IEEE Transactions on Pattern Analysis and
  Machine Intelligence}} \bibinfo{volume}{45}, \bibinfo{number}{3}
  (\bibinfo{year}{2022}), \bibinfo{pages}{3292--3310}.
\newblock


\bibitem[Luiten et~al\mbox{.}(2023)]%
        {luiten2023dynamic}
\bibfield{author}{\bibinfo{person}{Jonathon Luiten}, \bibinfo{person}{Georgios
  Kopanas}, \bibinfo{person}{Bastian Leibe}, {and} \bibinfo{person}{Deva
  Ramanan}.} \bibinfo{year}{2023}\natexlab{}.
\newblock \showarticletitle{Dynamic 3d gaussians: Tracking by persistent
  dynamic view synthesis}.
\newblock \bibinfo{journal}{\emph{arXiv preprint arXiv:2308.09713}}
  (\bibinfo{year}{2023}).
\newblock


\bibitem[Mallick et~al\mbox{.}(2024)]%
        {mallick2024taming}
\bibfield{author}{\bibinfo{person}{Saswat~Subhajyoti Mallick},
  \bibinfo{person}{Rahul Goel}, \bibinfo{person}{Bernhard Kerbl},
  \bibinfo{person}{Markus Steinberger}, \bibinfo{person}{Francisco~Vicente
  Carrasco}, {and} \bibinfo{person}{Fernando De~La~Torre}.}
  \bibinfo{year}{2024}\natexlab{}.
\newblock \showarticletitle{Taming 3dgs: High-quality radiance fields with
  limited resources}. In \bibinfo{booktitle}{\emph{SIGGRAPH Asia 2024
  Conference Papers}}. \bibinfo{pages}{1--11}.
\newblock


\bibitem[Mantiuk et~al\mbox{.}(2022)]%
        {mantiuk2022stelacsf}
\bibfield{author}{\bibinfo{person}{Rafa{\l}~K Mantiuk}, \bibinfo{person}{Maliha
  Ashraf}, {and} \bibinfo{person}{Alexandre Chapiro}.}
  \bibinfo{year}{2022}\natexlab{}.
\newblock \showarticletitle{stelaCSF: A unified model of contrast sensitivity
  as the function of spatio-temporal frequency, eccentricity, luminance and
  area}.
\newblock \bibinfo{journal}{\emph{ACM Transactions on Graphics (TOG)}}
  \bibinfo{volume}{41}, \bibinfo{number}{4} (\bibinfo{year}{2022}),
  \bibinfo{pages}{1--16}.
\newblock


\bibitem[Mantiuk et~al\mbox{.}(2021)]%
        {mantiuk2021fovvideovdp}
\bibfield{author}{\bibinfo{person}{Rafa{\l}~K Mantiuk}, \bibinfo{person}{Gyorgy
  Denes}, \bibinfo{person}{Alexandre Chapiro}, \bibinfo{person}{Anton
  Kaplanyan}, \bibinfo{person}{Gizem Rufo}, \bibinfo{person}{Romain Bachy},
  \bibinfo{person}{Trisha Lian}, {and} \bibinfo{person}{Anjul Patney}.}
  \bibinfo{year}{2021}\natexlab{}.
\newblock \showarticletitle{Fovvideovdp: A visible difference predictor for
  wide field-of-view video}.
\newblock \bibinfo{journal}{\emph{ACM Transactions on Graphics (TOG)}}
  \bibinfo{volume}{40}, \bibinfo{number}{4} (\bibinfo{year}{2021}),
  \bibinfo{pages}{1--19}.
\newblock


\bibitem[Meng et~al\mbox{.}(2020b)]%
        {meng20203d}
\bibfield{author}{\bibinfo{person}{Xiaoxu Meng}, \bibinfo{person}{Ruofei Du},
  \bibinfo{person}{Joseph~F JaJa}, {and} \bibinfo{person}{Amitabh Varshney}.}
  \bibinfo{year}{2020}\natexlab{b}.
\newblock \showarticletitle{3D-kernel foveated rendering for light fields}.
\newblock \bibinfo{journal}{\emph{IEEE Transactions on Visualization and
  Computer Graphics}} \bibinfo{volume}{27}, \bibinfo{number}{8}
  (\bibinfo{year}{2020}), \bibinfo{pages}{3350--3360}.
\newblock


\bibitem[Meng et~al\mbox{.}(2020a)]%
        {meng2020eye}
\bibfield{author}{\bibinfo{person}{Xiaoxu Meng}, \bibinfo{person}{Ruofei Du},
  {and} \bibinfo{person}{Amitabh Varshney}.} \bibinfo{year}{2020}\natexlab{a}.
\newblock \showarticletitle{Eye-dominance-guided foveated rendering}.
\newblock \bibinfo{journal}{\emph{IEEE transactions on visualization and
  computer graphics}} \bibinfo{volume}{26}, \bibinfo{number}{5}
  (\bibinfo{year}{2020}), \bibinfo{pages}{1972--1980}.
\newblock


\bibitem[Meng et~al\mbox{.}(2018)]%
        {meng2018kernel}
\bibfield{author}{\bibinfo{person}{Xiaoxu Meng}, \bibinfo{person}{Ruofei Du},
  \bibinfo{person}{Matthias Zwicker}, {and} \bibinfo{person}{Amitabh
  Varshney}.} \bibinfo{year}{2018}\natexlab{}.
\newblock \showarticletitle{{Kernel Foveated Rendering}}.
\newblock \bibinfo{journal}{\emph{Proceedings of the ACM on Computer Graphics
  and Interactive Techniques}} \bibinfo{volume}{1}, \bibinfo{number}{1}
  (\bibinfo{year}{2018}), \bibinfo{pages}{5}.
\newblock


\bibitem[Meyer et~al\mbox{.}(2024)]%
        {meyer2024pegasus}
\bibfield{author}{\bibinfo{person}{Lukas Meyer}, \bibinfo{person}{Floris
  Erich}, \bibinfo{person}{Yusuke Yoshiyasu}, \bibinfo{person}{Marc
  Stamminger}, \bibinfo{person}{Noriaki Ando}, {and} \bibinfo{person}{Yukiyasu
  Domae}.} \bibinfo{year}{2024}\natexlab{}.
\newblock \showarticletitle{PEGASUS: Physically Enhanced Gaussian Splatting
  Simulation System for 6DOF Object Pose Dataset Generation}.
\newblock \bibinfo{journal}{\emph{arXiv preprint arXiv:2401.02281}}
  (\bibinfo{year}{2024}).
\newblock


\bibitem[Mi and Xu(2023)]%
        {mi2023switchnerf}
\bibfield{author}{\bibinfo{person}{Zhenxing Mi} {and} \bibinfo{person}{Dan
  Xu}.} \bibinfo{year}{2023}\natexlab{}.
\newblock \showarticletitle{Switch-NeRF: Learning Scene Decomposition with
  Mixture of Experts for Large-scale Neural Radiance Fields}. In
  \bibinfo{booktitle}{\emph{International Conference on Learning
  Representations (ICLR)}}.
\newblock
\urldef\tempurl%
\url{https://openreview.net/forum?id=PQ2zoIZqvm}
\showURL{%
\tempurl}


\bibitem[Mildenhall et~al\mbox{.}(2020)]%
        {mildenhall2020nerf}
\bibfield{author}{\bibinfo{person}{Ben Mildenhall}, \bibinfo{person}{Pratul~P.
  Srinivasan}, \bibinfo{person}{Matthew Tancik}, \bibinfo{person}{Jonathan~T.
  Barron}, \bibinfo{person}{Ravi Ramamoorthi}, {and} \bibinfo{person}{Ren Ng}.}
  \bibinfo{year}{2020}\natexlab{}.
\newblock \showarticletitle{{NeRF}: Representing Scenes as Neural Radiance
  Fields for View Synthesis}. In \bibinfo{booktitle}{\emph{ECCV}}.
\newblock
\urldef\tempurl%
\url{https://doi.org/10.1145/3503250}
\showDOI{\tempurl}


\bibitem[M\"{u}ller et~al\mbox{.}(2022)]%
        {mueller2022instant}
\bibfield{author}{\bibinfo{person}{Thomas M\"{u}ller}, \bibinfo{person}{Alex
  Evans}, \bibinfo{person}{Christoph Schied}, {and} \bibinfo{person}{Alexander
  Keller}.} \bibinfo{year}{2022}\natexlab{}.
\newblock \showarticletitle{Instant neural graphics primitives with a
  multiresolution hash encoding}.
\newblock \bibinfo{journal}{\emph{ACM TOG}} \bibinfo{volume}{41},
  \bibinfo{number}{4}, Article \bibinfo{articleno}{102} (\bibinfo{date}{jul}
  \bibinfo{year}{2022}), \bibinfo{numpages}{15}~pages.
\newblock
\showISSN{0730-0301}
\urldef\tempurl%
\url{https://doi.org/10.1145/3528223.3530127}
\showDOI{\tempurl}


\bibitem[Neff et~al\mbox{.}(2021)]%
        {neff2021donerf}
\bibfield{author}{\bibinfo{person}{Thomas Neff}, \bibinfo{person}{Pascal
  Stadlbauer}, \bibinfo{person}{Mathias Parger}, \bibinfo{person}{Andreas
  Kurz}, \bibinfo{person}{Joerg~H. Mueller}, \bibinfo{person}{Chakravarty
  R.~Alla Chaitanya}, \bibinfo{person}{Anton~S. Kaplanyan}, {and}
  \bibinfo{person}{Markus Steinberger}.} \bibinfo{year}{2021}\natexlab{}.
\newblock \showarticletitle{{DONeRF: Towards Real-Time Rendering of Compact
  Neural Radiance Fields using Depth Oracle Networks}}.
\newblock \bibinfo{journal}{\emph{CGF}} (\bibinfo{year}{2021}).
\newblock
\urldef\tempurl%
\url{https://doi.org/10.1111/cgf.14340}
\showDOI{\tempurl}


\bibitem[Ost et~al\mbox{.}(2022)]%
        {ost2022neural}
\bibfield{author}{\bibinfo{person}{Julian Ost}, \bibinfo{person}{Issam
  Laradji}, \bibinfo{person}{Alejandro Newell}, \bibinfo{person}{Yuval Bahat},
  {and} \bibinfo{person}{Felix Heide}.} \bibinfo{year}{2022}\natexlab{}.
\newblock \showarticletitle{Neural point light fields}. In
  \bibinfo{booktitle}{\emph{Proceedings of the IEEE/CVF Conference on Computer
  Vision and Pattern Recognition}}. \bibinfo{pages}{18419--18429}.
\newblock


\bibitem[Patas(2023)]%
        {patas23gscuda}
\bibfield{author}{\bibinfo{person}{Janusch Patas}.}
  \bibinfo{year}{2023}\natexlab{}.
\newblock \bibinfo{title}{Gaussian Splatting Cuda}.
\newblock
  \bibinfo{howpublished}{\url{https://github.com/MrNeRF/gaussian-splatting-cuda}}.
\newblock


\bibitem[Patney et~al\mbox{.}(2016)]%
        {patney2016towards}
\bibfield{author}{\bibinfo{person}{Anjul Patney}, \bibinfo{person}{Marco
  Salvi}, \bibinfo{person}{Joohwan Kim}, \bibinfo{person}{Anton Kaplanyan},
  \bibinfo{person}{Chris Wyman}, \bibinfo{person}{Nir Benty},
  \bibinfo{person}{David Luebke}, {and} \bibinfo{person}{Aaron Lefohn}.}
  \bibinfo{year}{2016}\natexlab{}.
\newblock \showarticletitle{{Towards Foveated Rendering for Gaze-Tracked
  Virtual Reality}}.
\newblock \bibinfo{journal}{\emph{ACM Transactions on Graphics (TOG)}}
  \bibinfo{volume}{35}, \bibinfo{number}{6} (\bibinfo{year}{2016}),
  \bibinfo{pages}{179}.
\newblock


\bibitem[Pfister et~al\mbox{.}(2000)]%
        {pfister2000surfels}
\bibfield{author}{\bibinfo{person}{Hanspeter Pfister},
  \bibinfo{person}{Matthias Zwicker}, \bibinfo{person}{Jeroen Van~Baar}, {and}
  \bibinfo{person}{Markus Gross}.} \bibinfo{year}{2000}\natexlab{}.
\newblock \showarticletitle{Surfels: Surface elements as rendering primitives}.
  In \bibinfo{booktitle}{\emph{Proceedings of the 27th annual conference on
  Computer graphics and interactive techniques}}. \bibinfo{pages}{335--342}.
\newblock


\bibitem[Philip and Deschaintre(2023)]%
        {philip2023floaters}
\bibfield{author}{\bibinfo{person}{Julien Philip} {and}
  \bibinfo{person}{Valentin Deschaintre}.} \bibinfo{year}{2023}\natexlab{}.
\newblock \showarticletitle{Floaters No More: Radiance Field Gradient Scaling
  for Improved Near-Camera Training}.
\newblock  (\bibinfo{year}{2023}).
\newblock


\bibitem[Radl et~al\mbox{.}(2024)]%
        {radl2024stopthepop}
\bibfield{author}{\bibinfo{person}{Lukas Radl}, \bibinfo{person}{Michael
  Steiner}, \bibinfo{person}{Mathias Parger}, \bibinfo{person}{Alexander
  Weinrauch}, \bibinfo{person}{Bernhard Kerbl}, {and} \bibinfo{person}{Markus
  Steinberger}.} \bibinfo{year}{2024}\natexlab{}.
\newblock \showarticletitle{Stopthepop: Sorted gaussian splatting for
  view-consistent real-time rendering}.
\newblock \bibinfo{journal}{\emph{ACM TOG}} \bibinfo{volume}{43},
  \bibinfo{number}{4} (\bibinfo{year}{2024}), \bibinfo{pages}{1--17}.
\newblock


\bibitem[Rakhimov et~al\mbox{.}(2022)]%
        {rakhimov2022npbgpp}
\bibfield{author}{\bibinfo{person}{Ruslan Rakhimov},
  \bibinfo{person}{Andrei-Timotei Ardelean}, \bibinfo{person}{Victor
  Lempitsky}, {and} \bibinfo{person}{Evgeny Burnaev}.}
  \bibinfo{year}{2022}\natexlab{}.
\newblock \showarticletitle{{NPBG$++$}: Accelerating Neural Point-Based
  Graphics}. In \bibinfo{booktitle}{\emph{CVPR}}.
\newblock
\urldef\tempurl%
\url{https://doi.org/10.1109/CVPR52688.2022.01550}
\showDOI{\tempurl}


\bibitem[Reiser et~al\mbox{.}(2024)]%
        {reiser2024binary}
\bibfield{author}{\bibinfo{person}{Christian Reiser}, \bibinfo{person}{Stephan
  Garbin}, \bibinfo{person}{Pratul~P Srinivasan}, \bibinfo{person}{Dor Verbin},
  \bibinfo{person}{Richard Szeliski}, \bibinfo{person}{Ben Mildenhall},
  \bibinfo{person}{Jonathan~T Barron}, \bibinfo{person}{Peter Hedman}, {and}
  \bibinfo{person}{Andreas Geiger}.} \bibinfo{year}{2024}\natexlab{}.
\newblock \showarticletitle{Binary Opacity Grids: Capturing Fine Geometric
  Detail for Mesh-Based View Synthesis}.
\newblock \bibinfo{journal}{\emph{arXiv preprint arXiv:2402.12377}}
  (\bibinfo{year}{2024}).
\newblock


\bibitem[Reiser et~al\mbox{.}(2023)]%
        {reiser2023merf}
\bibfield{author}{\bibinfo{person}{Christian Reiser}, \bibinfo{person}{Richard
  Szeliski}, \bibinfo{person}{Dor Verbin}, \bibinfo{person}{Pratul~P.
  Srinivasan}, \bibinfo{person}{Ben Mildenhall}, \bibinfo{person}{Andreas
  Geiger}, \bibinfo{person}{Jonathan~T. Barron}, {and} \bibinfo{person}{Peter
  Hedman}.} \bibinfo{year}{2023}\natexlab{}.
\newblock \showarticletitle{MERF: Memory-Efficient Radiance Fields for
  Real-time View Synthesis in Unbounded Scenes}.
\newblock \bibinfo{journal}{\emph{SIGGRAPH}} (\bibinfo{year}{2023}).
\newblock
\urldef\tempurl%
\url{https://doi.org/10.1145/3592426}
\showDOI{\tempurl}


\bibitem[Ren et~al\mbox{.}(2024)]%
        {ren2024octreegs}
\bibfield{author}{\bibinfo{person}{Kerui Ren}, \bibinfo{person}{Lihan Jiang},
  \bibinfo{person}{Tao Lu}, \bibinfo{person}{Mulin Yu},
  \bibinfo{person}{Linning Xu}, \bibinfo{person}{Zhangkai Ni}, {and}
  \bibinfo{person}{Bo Dai}.} \bibinfo{year}{2024}\natexlab{}.
\newblock \bibinfo{title}{Octree-GS: Towards Consistent Real-time Rendering
  with LOD-Structured 3D Gaussians}.
\newblock
\newblock
\showeprint[arxiv]{2403.17898}~[cs.CV]


\bibitem[Rolff et~al\mbox{.}(2023a)]%
        {rolff2023interactive}
\bibfield{author}{\bibinfo{person}{Tim Rolff}, \bibinfo{person}{Ke Li},
  \bibinfo{person}{Julia Hertel}, \bibinfo{person}{Susanne Schmidt},
  \bibinfo{person}{Simone Frintrop}, {and} \bibinfo{person}{Frank Steinicke}.}
  \bibinfo{year}{2023}\natexlab{a}.
\newblock \showarticletitle{Interactive VRS-NeRF: Lightning fast Neural
  Radiance Field Rendering for Virtual Reality}. In
  \bibinfo{booktitle}{\emph{Proceedings of the 2023 ACM Symposium on Spatial
  User Interaction}}. \bibinfo{pages}{1--3}.
\newblock


\bibitem[Rolff et~al\mbox{.}(2023b)]%
        {rolff2023vrs}
\bibfield{author}{\bibinfo{person}{Tim Rolff}, \bibinfo{person}{Susanne
  Schmidt}, \bibinfo{person}{Ke Li}, \bibinfo{person}{Frank Steinicke}, {and}
  \bibinfo{person}{Simone Frintrop}.} \bibinfo{year}{2023}\natexlab{b}.
\newblock \showarticletitle{VRS-NeRF: Accelerating Neural Radiance Field
  Rendering with Variable Rate Shading}. In \bibinfo{booktitle}{\emph{2023 IEEE
  International Symposium on Mixed and Augmented Reality (ISMAR)}}. IEEE,
  \bibinfo{pages}{243--252}.
\newblock


\bibitem[R{\"u}ckert et~al\mbox{.}(2022)]%
        {ruckert2022adop}
\bibfield{author}{\bibinfo{person}{Darius R{\"u}ckert}, \bibinfo{person}{Linus
  Franke}, {and} \bibinfo{person}{Marc Stamminger}.}
  \bibinfo{year}{2022}\natexlab{}.
\newblock \showarticletitle{Adop: Approximate differentiable one-pixel point
  rendering}.
\newblock \bibinfo{journal}{\emph{ACM TOG}} (\bibinfo{year}{2022}).
\newblock
\urldef\tempurl%
\url{https://doi.org/10.1145/3528223.3530122}
\showDOI{\tempurl}


\bibitem[R\"{u}ckert et~al\mbox{.}(2022)]%
        {ruckert2022neat}
\bibfield{author}{\bibinfo{person}{Darius R\"{u}ckert},
  \bibinfo{person}{Yuanhao Wang}, \bibinfo{person}{Rui Li},
  \bibinfo{person}{Ramzi Idoughi}, {and} \bibinfo{person}{Wolfgang Heidrich}.}
  \bibinfo{year}{2022}\natexlab{}.
\newblock \showarticletitle{NeAT: Neural Adaptive Tomography}.
\newblock \bibinfo{journal}{\emph{ACM Trans. Graph.}} \bibinfo{volume}{41},
  \bibinfo{number}{4}, Article \bibinfo{articleno}{55} (\bibinfo{date}{jul}
  \bibinfo{year}{2022}), \bibinfo{numpages}{13}~pages.
\newblock
\showISSN{0730-0301}
\urldef\tempurl%
\url{https://doi.org/10.1145/3528223.3530121}
\showDOI{\tempurl}


\bibitem[Sch\"{o}nberger and Frahm(2016)]%
        {schoenberger2016sfm}
\bibfield{author}{\bibinfo{person}{Johannes~Lutz Sch\"{o}nberger} {and}
  \bibinfo{person}{Jan-Michael Frahm}.} \bibinfo{year}{2016}\natexlab{}.
\newblock \showarticletitle{Structure-from-Motion Revisited}. In
  \bibinfo{booktitle}{\emph{CVPR}}. \bibinfo{pages}{4104--4113}.
\newblock
\urldef\tempurl%
\url{https://doi.org/10.1109/CVPR.2016.445}
\showDOI{\tempurl}


\bibitem[Sch\"{o}nberger et~al\mbox{.}(2016)]%
        {schoenberger2016MVS}
\bibfield{author}{\bibinfo{person}{Johannes~Lutz Sch\"{o}nberger},
  \bibinfo{person}{Enliang Zheng}, \bibinfo{person}{Marc Pollefeys}, {and}
  \bibinfo{person}{Jan-Michael Frahm}.} \bibinfo{year}{2016}\natexlab{}.
\newblock \showarticletitle{Pixelwise View Selection for Unstructured
  Multi-View Stereo}. In \bibinfo{booktitle}{\emph{European Conference on
  Computer Vision (ECCV)}}.
\newblock


\bibitem[Sch{\"u}tz et~al\mbox{.}(2021)]%
        {schutz2021rendering}
\bibfield{author}{\bibinfo{person}{Markus Sch{\"u}tz},
  \bibinfo{person}{Bernhard Kerbl}, {and} \bibinfo{person}{Michael Wimmer}.}
  \bibinfo{year}{2021}\natexlab{}.
\newblock \showarticletitle{Rendering point clouds with compute shaders and
  vertex order optimization}. In \bibinfo{booktitle}{\emph{Computer Graphics
  Forum}}, Vol.~\bibinfo{volume}{40}. Wiley Online Library,
  \bibinfo{pages}{115--126}.
\newblock


\bibitem[Sch{\"u}tz et~al\mbox{.}(2022)]%
        {schutz2022software}
\bibfield{author}{\bibinfo{person}{Markus Sch{\"u}tz},
  \bibinfo{person}{Bernhard Kerbl}, {and} \bibinfo{person}{Michael Wimmer}.}
  \bibinfo{year}{2022}\natexlab{}.
\newblock \showarticletitle{Software rasterization of 2 billion points in real
  time}.
\newblock \bibinfo{journal}{\emph{ACM Comput. Graph. Int. Techn.}}
  \bibinfo{volume}{5}, \bibinfo{number}{3} (\bibinfo{year}{2022}),
  \bibinfo{pages}{1--17}.
\newblock


\bibitem[Sch{\"u}tz et~al\mbox{.}(2019)]%
        {schutz2019real}
\bibfield{author}{\bibinfo{person}{Markus Sch{\"u}tz},
  \bibinfo{person}{Katharina Kr{\"o}sl}, {and} \bibinfo{person}{Michael
  Wimmer}.} \bibinfo{year}{2019}\natexlab{}.
\newblock \showarticletitle{Real-time continuous level of detail rendering of
  point clouds}. In \bibinfo{booktitle}{\emph{2019 IEEE Conference on Virtual
  Reality and 3D User Interfaces (VR)}}. IEEE, \bibinfo{pages}{103--110}.
\newblock


\bibitem[Seitz et~al\mbox{.}(2006)]%
        {seitz2006comparison}
\bibfield{author}{\bibinfo{person}{Steven~M Seitz}, \bibinfo{person}{Brian
  Curless}, \bibinfo{person}{James Diebel}, \bibinfo{person}{Daniel
  Scharstein}, {and} \bibinfo{person}{Richard Szeliski}.}
  \bibinfo{year}{2006}\natexlab{}.
\newblock \showarticletitle{A comparison and evaluation of multi-view stereo
  reconstruction algorithms}. In \bibinfo{booktitle}{\emph{2006 IEEE computer
  society conference on computer vision and pattern recognition (CVPR'06)}},
  Vol.~\bibinfo{volume}{1}. IEEE, \bibinfo{pages}{519--528}.
\newblock


\bibitem[Shi et~al\mbox{.}(2024)]%
        {shi2024scene}
\bibfield{author}{\bibinfo{person}{Xuehuai Shi}, \bibinfo{person}{Lili Wang},
  \bibinfo{person}{Xinda Liu}, \bibinfo{person}{Jian Wu}, {and}
  \bibinfo{person}{Zhiwen Shao}.} \bibinfo{year}{2024}\natexlab{}.
\newblock \showarticletitle{Scene-aware Foveated Neural Radiance Fields}.
\newblock \bibinfo{journal}{\emph{IEEE Transactions on Visualization and
  Computer Graphics}} (\bibinfo{year}{2024}), \bibinfo{pages}{1--14}.
\newblock


\bibitem[Shum and Kang(2000)]%
        {shum2000review}
\bibfield{author}{\bibinfo{person}{Harry Shum} {and} \bibinfo{person}{Sing~Bing
  Kang}.} \bibinfo{year}{2000}\natexlab{}.
\newblock \showarticletitle{Review of image-based rendering techniques}. In
  \bibinfo{booktitle}{\emph{Visual Communications and Image Processing 2000}},
  Vol.~\bibinfo{volume}{4067}. SPIE, \bibinfo{pages}{2--13}.
\newblock


\bibitem[Sitzmann et~al\mbox{.}(2018)]%
        {sitzmann2018saliency}
\bibfield{author}{\bibinfo{person}{Vincent Sitzmann}, \bibinfo{person}{Ana
  Serrano}, \bibinfo{person}{Amy Pavel}, \bibinfo{person}{Maneesh Agrawala},
  \bibinfo{person}{Diego Gutierrez}, \bibinfo{person}{Belen Masia}, {and}
  \bibinfo{person}{Gordon Wetzstein}.} \bibinfo{year}{2018}\natexlab{}.
\newblock \showarticletitle{{Saliency in VR: How do people explore virtual
  environments?}}
\newblock \bibinfo{journal}{\emph{IEEE transactions on visualization and
  computer graphics}} \bibinfo{volume}{24}, \bibinfo{number}{4}
  (\bibinfo{year}{2018}), \bibinfo{pages}{1633--1642}.
\newblock


\bibitem[Snavely et~al\mbox{.}(2006)]%
        {snavely2006photo}
\bibfield{author}{\bibinfo{person}{Noah Snavely}, \bibinfo{person}{Steven~M
  Seitz}, {and} \bibinfo{person}{Richard Szeliski}.}
  \bibinfo{year}{2006}\natexlab{}.
\newblock \showarticletitle{Photo tourism: exploring photo collections in 3D}.
  In \bibinfo{booktitle}{\emph{ACM Siggraph 2006}}. \bibinfo{pages}{835--846}.
\newblock


\bibitem[Stengel et~al\mbox{.}(2016)]%
        {stengel2016adaptive}
\bibfield{author}{\bibinfo{person}{Michael Stengel}, \bibinfo{person}{Steve
  Grogorick}, \bibinfo{person}{Martin Eisemann}, {and} \bibinfo{person}{Marcus
  Magnor}.} \bibinfo{year}{2016}\natexlab{}.
\newblock \showarticletitle{{Adaptive Image-Space Sampling for Gaze-Contingent
  Real-Time Rendering}}. In \bibinfo{booktitle}{\emph{Computer Graphics
  Forum}}, Vol.~\bibinfo{volume}{35}. Wiley Online Library,
  \bibinfo{pages}{129--139}.
\newblock


\bibitem[Sun et~al\mbox{.}(2017)]%
        {sun2017perceptually}
\bibfield{author}{\bibinfo{person}{Qi Sun}, \bibinfo{person}{Fu-Chung Huang},
  \bibinfo{person}{Joohwan Kim}, \bibinfo{person}{Li-Yi Wei},
  \bibinfo{person}{David Luebke}, {and} \bibinfo{person}{Arie Kaufman}.}
  \bibinfo{year}{2017}\natexlab{}.
\newblock \showarticletitle{{Perceptually-guided foveation for light field
  displays}}.
\newblock \bibinfo{journal}{\emph{ACM Transactions on Graphics (TOG)}}
  \bibinfo{volume}{36}, \bibinfo{number}{6} (\bibinfo{year}{2017}),
  \bibinfo{pages}{192}.
\newblock


\bibitem[Swafford et~al\mbox{.}(2016)]%
        {swafford2016user}
\bibfield{author}{\bibinfo{person}{Nicholas~T Swafford},
  \bibinfo{person}{Jos{\'e}~A Iglesias-Guitian}, \bibinfo{person}{Charalampos
  Koniaris}, \bibinfo{person}{Bochang Moon}, \bibinfo{person}{Darren Cosker},
  {and} \bibinfo{person}{Kenny Mitchell}.} \bibinfo{year}{2016}\natexlab{}.
\newblock \showarticletitle{User, metric, and computational evaluation of
  foveated rendering methods}. In \bibinfo{booktitle}{\emph{Proceedings of the
  ACM Symposium on Applied Perception}}. \bibinfo{pages}{7--14}.
\newblock


\bibitem[Tancik et~al\mbox{.}(2022)]%
        {tancik2022block}
\bibfield{author}{\bibinfo{person}{Matthew Tancik}, \bibinfo{person}{Vincent
  Casser}, \bibinfo{person}{Xinchen Yan}, \bibinfo{person}{Sabeek Pradhan},
  \bibinfo{person}{Ben Mildenhall}, \bibinfo{person}{Pratul~P Srinivasan},
  \bibinfo{person}{Jonathan~T Barron}, {and} \bibinfo{person}{Henrik
  Kretzschmar}.} \bibinfo{year}{2022}\natexlab{}.
\newblock \showarticletitle{Block-NeRF: Scalable large scene neural view
  synthesis}. In \bibinfo{booktitle}{\emph{CVPR}}.
\newblock


\bibitem[Tancik et~al\mbox{.}(2021)]%
        {tancik2021learned}
\bibfield{author}{\bibinfo{person}{Matthew Tancik}, \bibinfo{person}{Ben
  Mildenhall}, \bibinfo{person}{Terrance Wang}, \bibinfo{person}{Divi Schmidt},
  \bibinfo{person}{Pratul~P Srinivasan}, \bibinfo{person}{Jonathan~T Barron},
  {and} \bibinfo{person}{Ren Ng}.} \bibinfo{year}{2021}\natexlab{}.
\newblock \showarticletitle{Learned initializations for optimizing
  coordinate-based neural representations}. In
  \bibinfo{booktitle}{\emph{CVPR}}. \bibinfo{pages}{2846--2855}.
\newblock


\bibitem[Tewari et~al\mbox{.}(2022)]%
        {Tewari2022NeuRendSTAR}
\bibfield{author}{\bibinfo{person}{A. Tewari}, \bibinfo{person}{J. Thies},
  \bibinfo{person}{B. Mildenhall}, \bibinfo{person}{P. Srinivasan},
  \bibinfo{person}{E. Tretschk}, \bibinfo{person}{W. Yifan},
  \bibinfo{person}{C. Lassner}, \bibinfo{person}{V. Sitzmann},
  \bibinfo{person}{R. Martin-Brualla}, \bibinfo{person}{S. Lombardi},
  \bibinfo{person}{T. Simon}, \bibinfo{person}{C. Theobalt},
  \bibinfo{person}{M. Nie{\ss}ner}, \bibinfo{person}{J.~T. Barron},
  \bibinfo{person}{G. Wetzstein}, \bibinfo{person}{M. Zollh{\"o}fer}, {and}
  \bibinfo{person}{V. Golyanik}.} \bibinfo{year}{2022}\natexlab{}.
\newblock \showarticletitle{{Advances in Neural Rendering}}.
\newblock \bibinfo{journal}{\emph{EG STAR}} (\bibinfo{year}{2022}).
\newblock
\urldef\tempurl%
\url{https://doi.org/10.1111/cgf.14507}
\showDOI{\tempurl}


\bibitem[Thies et~al\mbox{.}(2019)]%
        {thies2019deferred}
\bibfield{author}{\bibinfo{person}{Justus Thies}, \bibinfo{person}{Michael
  Zollh{\"o}fer}, {and} \bibinfo{person}{Matthias Nie{\ss}ner}.}
  \bibinfo{year}{2019}\natexlab{}.
\newblock \showarticletitle{Deferred neural rendering: Image synthesis using
  neural textures}.
\newblock \bibinfo{journal}{\emph{ACM TOG}} (\bibinfo{year}{2019}).
\newblock


\bibitem[Turki et~al\mbox{.}(2024)]%
        {turki2024hybridnerf}
\bibfield{author}{\bibinfo{person}{Haithem Turki}, \bibinfo{person}{Vasu
  Agrawal}, \bibinfo{person}{Samuel~Rota Bul{\`o}}, \bibinfo{person}{Lorenzo
  Porzi}, \bibinfo{person}{Peter Kontschieder}, \bibinfo{person}{Deva Ramanan},
  \bibinfo{person}{Michael Zollh{\"o}fer}, {and} \bibinfo{person}{Christian
  Richardt}.} \bibinfo{year}{2024}\natexlab{}.
\newblock \showarticletitle{HybridNeRF: Efficient Neural Rendering via Adaptive
  Volumetric Surfaces}. In \bibinfo{booktitle}{\emph{Proceedings of the
  IEEE/CVF Conference on Computer Vision and Pattern Recognition}}.
  \bibinfo{pages}{19647--19656}.
\newblock


\bibitem[Turki et~al\mbox{.}(2022)]%
        {turki2022mega}
\bibfield{author}{\bibinfo{person}{Haithem Turki}, \bibinfo{person}{Deva
  Ramanan}, {and} \bibinfo{person}{Mahadev Satyanarayanan}.}
  \bibinfo{year}{2022}\natexlab{}.
\newblock \showarticletitle{Mega-NeRF: Scalable Construction of Large-Scale
  NeRFs for Virtual Fly-Throughs}. In \bibinfo{booktitle}{\emph{CVPR}}.
  \bibinfo{pages}{12922--12931}.
\newblock


\bibitem[Tursun et~al\mbox{.}(2019)]%
        {tursun2019luminance}
\bibfield{author}{\bibinfo{person}{Okan~Tarhan Tursun}, \bibinfo{person}{Elena
  Arabadzhiyska-Koleva}, \bibinfo{person}{Marek Wernikowski},
  \bibinfo{person}{Rados{\l}aw Mantiuk}, \bibinfo{person}{Hans-Peter Seidel},
  \bibinfo{person}{Karol Myszkowski}, {and} \bibinfo{person}{Piotr Didyk}.}
  \bibinfo{year}{2019}\natexlab{}.
\newblock \showarticletitle{Luminance-contrast-aware foveated rendering}.
\newblock \bibinfo{journal}{\emph{ACM Transactions on Graphics (TOG)}}
  \bibinfo{volume}{38}, \bibinfo{number}{4} (\bibinfo{year}{2019}),
  \bibinfo{pages}{1--14}.
\newblock


\bibitem[Wang et~al\mbox{.}(2023)]%
        {wang2023foveated}
\bibfield{author}{\bibinfo{person}{Lili Wang}, \bibinfo{person}{Xuehuai Shi},
  {and} \bibinfo{person}{Yi Liu}.} \bibinfo{year}{2023}\natexlab{}.
\newblock \showarticletitle{Foveated rendering: A state-of-the-art survey}.
\newblock \bibinfo{journal}{\emph{Computational Visual Media}}
  \bibinfo{volume}{9}, \bibinfo{number}{2} (\bibinfo{year}{2023}),
  \bibinfo{pages}{195--228}.
\newblock


\bibitem[Wang et~al\mbox{.}(2024)]%
        {wang2024vprf}
\bibfield{author}{\bibinfo{person}{Zijun Wang}, \bibinfo{person}{Jian Wu},
  \bibinfo{person}{Runze Fan}, \bibinfo{person}{Wei Ke}, {and}
  \bibinfo{person}{Lili Wang}.} \bibinfo{year}{2024}\natexlab{}.
\newblock \showarticletitle{VPRF: Visual Perceptual Radiance Fields for
  Foveated Image Synthesis}.
\newblock \bibinfo{journal}{\emph{IEEE Transactions on Visualization and
  Computer Graphics}} (\bibinfo{year}{2024}).
\newblock


\bibitem[Weier et~al\mbox{.}(2018)]%
        {weier2018foveated}
\bibfield{author}{\bibinfo{person}{Martin Weier}, \bibinfo{person}{Thorsten
  Roth}, \bibinfo{person}{Andr{\'e} Hinkenjann}, {and} \bibinfo{person}{Philipp
  Slusallek}.} \bibinfo{year}{2018}\natexlab{}.
\newblock \showarticletitle{{Foveated Depth-of-Field Filtering in Head-Mounted
  Displays}}.
\newblock \bibinfo{journal}{\emph{ACM Transactions on Applied Perception
  (TAP)}} \bibinfo{volume}{15}, \bibinfo{number}{4} (\bibinfo{year}{2018}),
  \bibinfo{pages}{26}.
\newblock


\bibitem[Weier et~al\mbox{.}(2016)]%
        {weier2016foveated}
\bibfield{author}{\bibinfo{person}{Martin Weier}, \bibinfo{person}{Thorsten
  Roth}, \bibinfo{person}{Ernst Kruijff}, \bibinfo{person}{Andr{\'e}
  Hinkenjann}, \bibinfo{person}{Ars{\`e}ne P{\'e}rard-Gayot},
  \bibinfo{person}{Philipp Slusallek}, {and} \bibinfo{person}{Yongmin Li}.}
  \bibinfo{year}{2016}\natexlab{}.
\newblock \showarticletitle{{Foveated Real-Time Ray Tracing for Head-Mounted
  Displays}}. In \bibinfo{booktitle}{\emph{Computer Graphics Forum}},
  Vol.~\bibinfo{volume}{35}. Wiley Online Library, \bibinfo{pages}{289--298}.
\newblock


\bibitem[Weier et~al\mbox{.}(2017)]%
        {weier2017perception}
\bibfield{author}{\bibinfo{person}{Martin Weier}, \bibinfo{person}{Michael
  Stengel}, \bibinfo{person}{Thorsten Roth}, \bibinfo{person}{Piotr Didyk},
  \bibinfo{person}{Elmar Eisemann}, \bibinfo{person}{Martin Eisemann},
  \bibinfo{person}{Steve Grogorick}, \bibinfo{person}{Andr{\'e} Hinkenjann},
  \bibinfo{person}{Ernst Kruijff}, \bibinfo{person}{Marcus Magnor},
  {et~al\mbox{.}}} \bibinfo{year}{2017}\natexlab{}.
\newblock \showarticletitle{{Perception-Driven Accelerated Rendering}}. In
  \bibinfo{booktitle}{\emph{Computer Graphics Forum}},
  Vol.~\bibinfo{volume}{36}. Wiley Online Library, \bibinfo{pages}{611--643}.
\newblock


\bibitem[Whelan et~al\mbox{.}(2016)]%
        {whelan2016elasticfusion}
\bibfield{author}{\bibinfo{person}{Thomas Whelan}, \bibinfo{person}{Renato~F
  Salas-Moreno}, \bibinfo{person}{Ben Glocker}, \bibinfo{person}{Andrew~J
  Davison}, {and} \bibinfo{person}{Stefan Leutenegger}.}
  \bibinfo{year}{2016}\natexlab{}.
\newblock \showarticletitle{ElasticFusion: Real-time dense SLAM and light
  source estimation}.
\newblock \bibinfo{journal}{\emph{The International Journal of Robotics
  Research}} \bibinfo{volume}{35}, \bibinfo{number}{14} (\bibinfo{year}{2016}),
  \bibinfo{pages}{1697--1716}.
\newblock


\bibitem[Wiles et~al\mbox{.}(2020)]%
        {wiles2020synsin}
\bibfield{author}{\bibinfo{person}{Olivia Wiles}, \bibinfo{person}{Georgia
  Gkioxari}, \bibinfo{person}{Richard Szeliski}, {and} \bibinfo{person}{Justin
  Johnson}.} \bibinfo{year}{2020}\natexlab{}.
\newblock \showarticletitle{SynSin: End-to-End View Synthesis From a Single
  Image}. In \bibinfo{booktitle}{\emph{CVPR}}.
\newblock
\urldef\tempurl%
\url{https://doi.org/10.1109/CVPR42600.2020.00749}
\showDOI{\tempurl}


\bibitem[Wu et~al\mbox{.}(2024a)]%
        {wu2024reconfusion}
\bibfield{author}{\bibinfo{person}{Rundi Wu}, \bibinfo{person}{Ben Mildenhall},
  \bibinfo{person}{Philipp Henzler}, \bibinfo{person}{Keunhong Park},
  \bibinfo{person}{Ruiqi Gao}, \bibinfo{person}{Daniel Watson},
  \bibinfo{person}{Pratul~P Srinivasan}, \bibinfo{person}{Dor Verbin},
  \bibinfo{person}{Jonathan~T Barron}, \bibinfo{person}{Ben Poole},
  {et~al\mbox{.}}} \bibinfo{year}{2024}\natexlab{a}.
\newblock \showarticletitle{Reconfusion: 3d reconstruction with diffusion
  priors}. In \bibinfo{booktitle}{\emph{CVPR}}. \bibinfo{pages}{21551--21561}.
\newblock


\bibitem[Wu et~al\mbox{.}(2024b)]%
        {wu2024recent}
\bibfield{author}{\bibinfo{person}{Tong Wu}, \bibinfo{person}{Yu-Jie Yuan},
  \bibinfo{person}{Ling-Xiao Zhang}, \bibinfo{person}{Jie Yang},
  \bibinfo{person}{Yan-Pei Cao}, \bibinfo{person}{Ling-Qi Yan}, {and}
  \bibinfo{person}{Lin Gao}.} \bibinfo{year}{2024}\natexlab{b}.
\newblock \showarticletitle{Recent advances in 3d gaussian splatting}.
\newblock \bibinfo{journal}{\emph{Computational Visual Media}}
  (\bibinfo{year}{2024}), \bibinfo{pages}{1--30}.
\newblock


\bibitem[Xu et~al\mbox{.}(2023)]%
        {xu2023vr}
\bibfield{author}{\bibinfo{person}{Linning Xu}, \bibinfo{person}{Vasu Agrawal},
  \bibinfo{person}{William Laney}, \bibinfo{person}{Tony Garcia},
  \bibinfo{person}{Aayush Bansal}, \bibinfo{person}{Changil Kim},
  \bibinfo{person}{Samuel Rota~Bul\`{o}}, \bibinfo{person}{Lorenzo Porzi},
  \bibinfo{person}{Peter Kontschieder}, \bibinfo{person}{Alja\v{z}
  Bo\v{z}i\v{c}}, \bibinfo{person}{Dahua Lin}, \bibinfo{person}{Michael
  Zollh\"{o}fer}, {and} \bibinfo{person}{Christian Richardt}.}
  \bibinfo{year}{2023}\natexlab{}.
\newblock \showarticletitle{VR-NeRF: High-Fidelity Virtualized Walkable
  Spaces}. In \bibinfo{booktitle}{\emph{SIGGRAPH Asia 2023 Conference Papers}}
  (Sydney, NSW, Australia) \emph{(\bibinfo{series}{SA '23})}.
  \bibinfo{publisher}{Association for Computing Machinery},
  \bibinfo{address}{New York, NY, USA}, Article \bibinfo{articleno}{43},
  \bibinfo{numpages}{12}~pages.
\newblock
\showISBNx{9798400703157}
\urldef\tempurl%
\url{https://doi.org/10.1145/3610548.3618139}
\showDOI{\tempurl}


\bibitem[Xu et~al\mbox{.}(2022)]%
        {xu2022point}
\bibfield{author}{\bibinfo{person}{Qiangeng Xu}, \bibinfo{person}{Zexiang Xu},
  \bibinfo{person}{Julien Philip}, \bibinfo{person}{Sai Bi},
  \bibinfo{person}{Zhixin Shu}, \bibinfo{person}{Kalyan Sunkavalli}, {and}
  \bibinfo{person}{Ulrich Neumann}.} \bibinfo{year}{2022}\natexlab{}.
\newblock \showarticletitle{Point-NeRF: Point-based Neural Radiance Fields}. In
  \bibinfo{booktitle}{\emph{CVPR}}.
\newblock
\urldef\tempurl%
\url{https://doi.org/10.1109/CVPR52688.2022.00536}
\showDOI{\tempurl}


\bibitem[Yang et~al\mbox{.}(2020)]%
        {Yang_2020_CVPR}
\bibfield{author}{\bibinfo{person}{Zhenpei Yang}, \bibinfo{person}{Yuning
  Chai}, \bibinfo{person}{Dragomir Anguelov}, \bibinfo{person}{Yin Zhou},
  \bibinfo{person}{Pei Sun}, \bibinfo{person}{Dumitru Erhan},
  \bibinfo{person}{Sean Rafferty}, {and} \bibinfo{person}{Henrik Kretzschmar}.}
  \bibinfo{year}{2020}\natexlab{}.
\newblock \showarticletitle{SurfelGAN: Synthesizing Realistic Sensor Data for
  Autonomous Driving}. In \bibinfo{booktitle}{\emph{Proceedings of the IEEE/CVF
  Conference on Computer Vision and Pattern Recognition (CVPR)}}.
\newblock


\bibitem[Yang et~al\mbox{.}(2023)]%
        {yang2023deformable}
\bibfield{author}{\bibinfo{person}{Ziyi Yang}, \bibinfo{person}{Xinyu Gao},
  \bibinfo{person}{Wen Zhou}, \bibinfo{person}{Shaohui Jiao},
  \bibinfo{person}{Yuqing Zhang}, {and} \bibinfo{person}{Xiaogang Jin}.}
  \bibinfo{year}{2023}\natexlab{}.
\newblock \showarticletitle{Deformable 3d gaussians for high-fidelity monocular
  dynamic scene reconstruction}.
\newblock \bibinfo{journal}{\emph{arXiv preprint arXiv:2309.13101}}
  (\bibinfo{year}{2023}).
\newblock


\bibitem[Yariv et~al\mbox{.}(2023)]%
        {yariv2023bakedsdf}
\bibfield{author}{\bibinfo{person}{Lior Yariv}, \bibinfo{person}{Peter Hedman},
  \bibinfo{person}{Christian Reiser}, \bibinfo{person}{Dor Verbin},
  \bibinfo{person}{Pratul~P. Srinivasan}, \bibinfo{person}{Richard Szeliski},
  \bibinfo{person}{Jonathan~T. Barron}, {and} \bibinfo{person}{Ben
  Mildenhall}.} \bibinfo{year}{2023}\natexlab{}.
\newblock \showarticletitle{{BakedSDF}: Meshing Neural {SDFs} for Real-Time
  View Synthesis}. In \bibinfo{booktitle}{\emph{SIGGRAPH}} (Los Angeles, CA,
  USA) \emph{(\bibinfo{series}{SIGGRAPH '23})}. \bibinfo{publisher}{Association
  for Computing Machinery}, Article \bibinfo{articleno}{46},
  \bibinfo{numpages}{9}~pages.
\newblock
\urldef\tempurl%
\url{https://doi.org/10.1145/3588432.3591536}
\showDOI{\tempurl}


\bibitem[Ye et~al\mbox{.}(2024b)]%
        {ye2024neural}
\bibfield{author}{\bibinfo{person}{Jiannan Ye}, \bibinfo{person}{Xiaoxu Meng},
  \bibinfo{person}{Daiyun Guo}, \bibinfo{person}{Cheng Shang},
  \bibinfo{person}{Haotian Mao}, {and} \bibinfo{person}{Xubo Yang}.}
  \bibinfo{year}{2024}\natexlab{b}.
\newblock \showarticletitle{Neural foveated super-resolution for real-time VR
  rendering}.
\newblock \bibinfo{journal}{\emph{Computer Animation and Virtual Worlds}}
  \bibinfo{volume}{35}, \bibinfo{number}{4} (\bibinfo{year}{2024}),
  \bibinfo{pages}{e2287}.
\newblock


\bibitem[Ye et~al\mbox{.}(2024a)]%
        {ye2024absgs}
\bibfield{author}{\bibinfo{person}{Zongxin Ye}, \bibinfo{person}{Wenyu Li},
  \bibinfo{person}{Sidun Liu}, \bibinfo{person}{Peng Qiao}, {and}
  \bibinfo{person}{Yong Dou}.} \bibinfo{year}{2024}\natexlab{a}.
\newblock \bibinfo{title}{AbsGS: Recovering Fine Details for 3D Gaussian
  Splatting}.
\newblock
\newblock
\showeprint[arxiv]{2404.10484}~[cs.CV]


\bibitem[Yifan et~al\mbox{.}(2019)]%
        {yifan2019differentiable}
\bibfield{author}{\bibinfo{person}{Wang Yifan}, \bibinfo{person}{Felice
  Serena}, \bibinfo{person}{Shihao Wu}, \bibinfo{person}{Cengiz {\"O}ztireli},
  {and} \bibinfo{person}{Olga Sorkine-Hornung}.}
  \bibinfo{year}{2019}\natexlab{}.
\newblock \showarticletitle{Differentiable surface splatting for point-based
  geometry processing}.
\newblock \bibinfo{journal}{\emph{ACM TOG}} \bibinfo{volume}{38},
  \bibinfo{number}{6} (\bibinfo{year}{2019}), \bibinfo{pages}{1--14}.
\newblock


\bibitem[Yu et~al\mbox{.}(2021a)]%
        {yu2021plenoctrees}
\bibfield{author}{\bibinfo{person}{Alex Yu}, \bibinfo{person}{Ruilong Li},
  \bibinfo{person}{Matthew Tancik}, \bibinfo{person}{Hao Li},
  \bibinfo{person}{Ren Ng}, {and} \bibinfo{person}{Angjoo Kanazawa}.}
  \bibinfo{year}{2021}\natexlab{a}.
\newblock \showarticletitle{{PlenOctrees} for Real-time Rendering of Neural
  Radiance Fields}. In \bibinfo{booktitle}{\emph{ICCV}}.
\newblock
\urldef\tempurl%
\url{https://doi.org/10.1109/ICCV48922.2021.00570}
\showDOI{\tempurl}


\bibitem[Yu et~al\mbox{.}(2021b)]%
        {yu2021pixelnerf}
\bibfield{author}{\bibinfo{person}{Alex Yu}, \bibinfo{person}{Vickie Ye},
  \bibinfo{person}{Matthew Tancik}, {and} \bibinfo{person}{Angjoo Kanazawa}.}
  \bibinfo{year}{2021}\natexlab{b}.
\newblock \showarticletitle{{pixelNeRF}: Neural radiance fields from one or few
  images}. In \bibinfo{booktitle}{\emph{CVPR}}. \bibinfo{pages}{4578--4587}.
\newblock
\urldef\tempurl%
\url{https://doi.org/10.1109/CVPR46437.2021.00455}
\showDOI{\tempurl}


\bibitem[Yu et~al\mbox{.}(2024)]%
        {yu2024mip}
\bibfield{author}{\bibinfo{person}{Zehao Yu}, \bibinfo{person}{Anpei Chen},
  \bibinfo{person}{Binbin Huang}, \bibinfo{person}{Torsten Sattler}, {and}
  \bibinfo{person}{Andreas Geiger}.} \bibinfo{year}{2024}\natexlab{}.
\newblock \showarticletitle{Mip-splatting: Alias-free 3d gaussian splatting}.
  In \bibinfo{booktitle}{\emph{Proceedings of the IEEE/CVF Conference on
  Computer Vision and Pattern Recognition}}. \bibinfo{pages}{19447--19456}.
\newblock


\bibitem[Zhang et~al\mbox{.}(2022)]%
        {zhang2022differentiable}
\bibfield{author}{\bibinfo{person}{Qiang Zhang}, \bibinfo{person}{Seung-Hwan
  Baek}, \bibinfo{person}{Szymon Rusinkiewicz}, {and} \bibinfo{person}{Felix
  Heide}.} \bibinfo{year}{2022}\natexlab{}.
\newblock \showarticletitle{Differentiable point-based radiance fields for
  efficient view synthesis}.
\newblock \bibinfo{journal}{\emph{arXiv preprint arXiv:2205.14330}}
  (\bibinfo{year}{2022}).
\newblock


\bibitem[Zhang et~al\mbox{.}(2018)]%
        {zhang2018lpips}
\bibfield{author}{\bibinfo{person}{Richard Zhang}, \bibinfo{person}{Phillip
  Isola}, \bibinfo{person}{Alexei~A Efros}, \bibinfo{person}{Eli Shechtman},
  {and} \bibinfo{person}{Oliver Wang}.} \bibinfo{year}{2018}\natexlab{}.
\newblock \showarticletitle{The unreasonable effectiveness of deep features as
  a perceptual metric}. In \bibinfo{booktitle}{\emph{CVPR}}.
\newblock
\urldef\tempurl%
\url{https://doi.org/10.1109/CVPR.2018.00068}
\showDOI{\tempurl}


\bibitem[Zhu et~al\mbox{.}(2024)]%
        {zhu2024rpbg}
\bibfield{author}{\bibinfo{person}{Qingtian Zhu}, \bibinfo{person}{Zizhuang
  Wei}, \bibinfo{person}{Zhongtian Zheng}, \bibinfo{person}{Yifan Zhan},
  \bibinfo{person}{Zhuyu Yao}, \bibinfo{person}{Jiawang Zhang},
  \bibinfo{person}{Kejian Wu}, {and} \bibinfo{person}{Yinqiang Zheng}.}
  \bibinfo{year}{2024}\natexlab{}.
\newblock \showarticletitle{RPBG: Towards Robust Neural Point-based Graphics in
  the Wild}.
\newblock \bibinfo{journal}{\emph{arXiv preprint arXiv:2405.05663}}
  (\bibinfo{year}{2024}).
\newblock


\bibitem[Zuo and Deng(2023)]%
        {zuo2022snp}
\bibfield{author}{\bibinfo{person}{Yiming Zuo} {and} \bibinfo{person}{Jia
  Deng}.} \bibinfo{year}{2023}\natexlab{}.
\newblock \showarticletitle{View Synthesis with Sculpted Neural Points}. In
  \bibinfo{booktitle}{\emph{ICLR}}, Vol.~\bibinfo{volume}{abs/2205.05869}.
\newblock
\urldef\tempurl%
\url{https://doi.org/10.48550/arXiv.2205.05869}
\showDOI{\tempurl}


\bibitem[Zwicker et~al\mbox{.}(2001)]%
        {zwicker2001surface}
\bibfield{author}{\bibinfo{person}{Matthias Zwicker},
  \bibinfo{person}{Hanspeter Pfister}, \bibinfo{person}{Jeroen Van~Baar}, {and}
  \bibinfo{person}{Markus Gross}.} \bibinfo{year}{2001}\natexlab{}.
\newblock \showarticletitle{Surface Splatting}. In
  \bibinfo{booktitle}{\emph{Proceedings of the 28th annual conference on
  Computer graphics and interactive techniques}}. \bibinfo{pages}{371--378}.
\newblock


\end{thebibliography}

\end{document}